\pdfoutput=1
\documentclass{article}
\usepackage{iclr2027_conference,times}

\usepackage{amsmath,amsfonts,bm}

\usepackage{hyperref}
\hypersetup{colorlinks=true, linkcolor=red!60!black, citecolor=blue!60!black, urlcolor=blue!60!black}
\usepackage{url}
\usepackage{graphicx}
\usepackage{booktabs}
\usepackage{xcolor}
\usepackage{etoolbox}
\usepackage{enumitem}
\usepackage{wrapfig}
\usepackage{placeins}
\AtBeginEnvironment{table}{%
  \setlength{\abovecaptionskip}{0pt}\setlength{\belowcaptionskip}{6pt}}

\newcommand{\methodlong}{Movement Prototype Quantization}

\newcommand{\proto}{movement prototype}
\newcommand{\protos}{movement prototypes}
\newcommand{\Proto}{Movement prototype}

\title{What Does a Skill Vocabulary Cost? \\ A Closed-Loop Decomposition over \\Demonstrated Motions}

\author{Maximilian Xiling Li$^{1}$, Nils Blank$^{1}$, Paul Mattes$^{1}$, \\
\textbf{Pankhuri Vanjani$^{1}$ \& Rudolf Lioutikov$^{1,2}$} \\
$^{1}$Intuitive Robots Lab, Karlsruhe Institute of Technology (KIT), Germany \\
$^{2}$Robotics Institute Germany (RIG) \\
\texttt{\{maximilian.li, lioutikov\}@kit.edu}
}

\iclrfinalcopy
\hypersetup{pdftitle={What Does a Skill Vocabulary Cost? A Closed-Loop Decomposition over Demonstrated Motions}}
\hypersetup{pdfauthor={Maximilian Xiling Li, Nils Blank, Paul Mattes, Pankhuri Vanjani, Rudolf Lioutikov}}
\begin{document}

\maketitle

\begin{abstract}
Discrete skill policies execute motions drawn from a finite vocabulary.
In most current methods, this vocabulary is a latent codebook learned
jointly with a selector and a decoder. When such a policy fails in
closed loop, the failure cannot be traced to the vocabulary, to the
choice of skill or to its execution. We present Movement Prototype
Quantization (MPQ), which places a library of demonstrated motions
after a trained policy instead of inside it. MPQ replaces each predicted
action chunk with its nearest library element, corrects that element
toward the prediction within a bound, and trains no parameters. Across
six architectures, including a 7B vision-language-action model, the
wrapped policy is statistically equivalent to the original. A reference
policy executed through the library is therefore a baseline for any
selector over the same library. Substituting a selector's components one
at a time attributes its shortfall to them. On LIBERO-90, with selectors
on three backbones, quantization costs nothing and the choice of skill
costs at most 2~points under the reference's placement and correction.
The selectors' own placement and correction account for nearly all of
the remaining gap, up to 12~points. Offline reconstruction predicts none
of these differences and misjudges library size, design and transfer
across benchmark suites. Project page: \url{https://intuitive-robots.github.io/mpq}.
\end{abstract}

\section{Introduction}
\label{sec:intro}

A policy that predicts action chunks can be given a finite vocabulary of motions
in one of two places. The vocabulary can sit inside the policy, as a latent
codebook learned jointly with a selector and a decoder \citep{lee2024vqbet,
mete2024quest, zheng2024prise}. It can also sit after the policy and be applied
to the predictions of a continuous policy, which then trains as usual while
every chunk it executes belongs to a finite set. Either placement allows a
pretrained vision-language model to emit actions as tokens
\citep{pertsch2025fast}, but the two differ in what the vocabulary
costs. This paper measures that cost.

\begin{figure}[t]
  \centering
  \includegraphics[width=\linewidth]{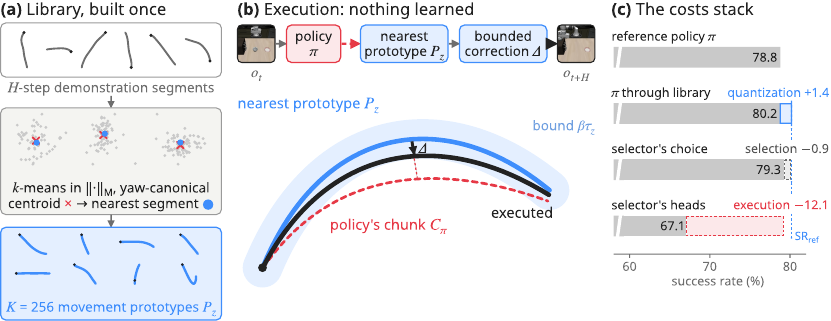}
  \caption{
    \methodlong. \textbf{(a)} Library construction. Demonstration segments are
    fit to B-spline control points, yaw-canonicalized, clustered under the Gram
    metric and projected onto real segments (LIBERO-90). \textbf{(b)} Execution.
    The chunk $C_\pi$ of a frozen policy is matched to its nearest \proto{}
    $P_z$, placed in the chunk's frame and corrected toward $C_\pi$ within the
    deviation bound (Eq.~\ref{eq:bound}). Blue: demonstrated. Red: predicted.
    Black: executed. \textbf{(c)} Decomposition. A reference policy executed
    through the library ($\mathrm{SR}_{\mathrm{ref}}$, dashed) sets the
    baseline. Substituting a selector's choice, then its
    placement and correction, splits the selector's shortfall into
    selection and execution (LIBERO-90, Table~\ref{tab:decomp}).
  }
  \label{fig:overview}
\end{figure}

Placing the vocabulary inside the policy has two consequences. First, a codebook
trained jointly with its selector and decoder entangles the costs of the
vocabulary, the selection and the execution, so closed-loop success reports only
their sum. Second, a code becomes a trajectory only through the learned decoder,
so the library is judged by offline reconstruction rather than by closed-loop
success.

We separate these costs by placing the vocabulary after the policy. If a
continuous reference policy keeps its success when executed through an explicit
library, the library covers the motions the task requires.
The reference executed through the library, the \emph{library reference},
  is then a baseline for any selector over that library.
A trained selector's shortfall against it can be split into the cost of
its choice and the cost of its placement and correction
(\S\ref{sec:instrument}). This requires two things. Every library element must
execute without a learned decoder, and the layer that applies the library must
cost the reference nothing.

Movement Prototype Quantization (MPQ) provides such a layer. Its elements, \emph{\protos{}}, are
demonstration segments stored as B-spline control points \citep{zhou2025beast}.
A fixed linear map decodes them, so every element is an executable trajectory.
At execution time, MPQ replaces each predicted chunk with its nearest
\proto{} and corrects it toward the prediction, within a bound set by the
demonstrations' own spread (Fig.~\ref{fig:overview}). Nothing on the execution
path is trained.

The layer costs the policies it wraps almost nothing (\S\ref{sec:exp}). Pooled
over 13 combinations of policy and benchmark on LIBERO \citep{liu2023libero} and
RoboCasa \citep{nasiriany2024robocasa}, covering six architectures including a
7B vision-language-action model, MPQ changes success by $+0.6$~points
(95\% CI $[-0.8, +1.9]$). Its cost does not grow under observation shift, and on
a real robot it runs with the bound selected in simulation.

Relative to the library reference,
selectors on three backbones lose almost all of their gap in placing and
  correcting the chosen element
(\S\ref{sec:placement}). On LIBERO-90 with 256 \protos{}, quantization costs
nothing. A selector's choice often differs from the nearest \proto{}, yet
with the reference's placement and correction it costs at
  most 2~points. The selectors' own placement and correction cost
  between 0 and 12~points.

Offline reconstruction, the standard criterion for validating a library,
predicts none of these differences (\S\ref{sec:libraries}). It keeps improving
after closed-loop success has saturated, it ranks library designs differently
than closed-loop success does, and it misses a transfer loss of 4 to 8~points.
As with policy checkpoints \citep{mandlekar2021matters}, offline
criteria therefore cannot select a library.

\paragraph{Contributions.}

\begin{itemize}[leftmargin=1.5em,itemsep=1pt,topsep=2pt,parsep=0pt]
  \item A closed-loop decomposition that splits a discrete skill policy's
        shortfall into quantization, selection and execution, measured against
        the library reference (\S\ref{sec:instrument}).
  \item MPQ, an execution layer that keeps every executed chunk within a
        bounded distance of a demonstrated segment and is statistically
        equivalent to the unwrapped policy, within the noise of repeated
        evaluation (\S\ref{sec:method}, \S\ref{sec:exp}).
  \item The finding that
        on three backbones, a selector's gap to the library reference
          arises in executing the chosen element rather than in the
          vocabulary or the choice
        (\S\ref{sec:placement}).
  \item Evidence that offline reconstruction misjudges library size, design and
        transfer (\S\ref{sec:libraries}).
\end{itemize}

\section{Related work}
\label{sec:related}

\paragraph{Discrete skill policies and how they are measured.}
VQ-BeT \citep{lee2024vqbet}, QueST \citep{mete2024quest} and PRISE
\citep{zheng2024prise} discretize action chunks into latent codebooks
learned as vector-quantized autoencoders \citep{oord2017vqvae}. FAST
\citep{pertsch2025fast} and its successors
\citep{liu2026oat,kang2026xtokenizer} instead tokenize chunks so that
autoregressive vision-language-action models can predict them. In the codebook
methods, a learned head selects the element, and a code becomes a
trajectory only through a learned decoder. The codebook is therefore validated offline, by reconstruction. In
imitation learning, \citet{mandlekar2021matters} showed that selecting
checkpoints by validation loss yields policies far below the best, and
evaluated them in closed loop instead. The library reference applies the same
principle to the vocabulary (\S\ref{sec:libraries}).

\paragraph{Selectors with learned offsets.}
BeT \citep{shafiullah2022bet} bins single-step actions and adds a
learned offset to the selected bin. This combination of bin and offset
makes it the closest structural relative of a selector over MPQ's
library. It differs from MPQ in three respects, however. Its bins
are means of raw single-step actions, its selector is a trained
classifier, and its offsets are unbounded. Concurrent work pairs a
learned selector with a residual that is free to overwrite the selected
code \citep{lei2026pfdag,wang2026arp}. For such a design,
Eq.~\ref{eq:decomp} measures the cost of the choice and the cost of the
heads separately. In MPQ the residual is bounded, so the executed
segment keeps the identity of the selected element.

\paragraph{Motion primitives and prototypes.}
Our construction, which clusters demonstration segments and projects the
centroids onto real ones, has classical relatives in libraries of motion
primitives \citep{ijspeert2013dynamical,paraschos2013probabilistic,li2023prodmp,loedige2025force}
and in skills segmented from demonstrations
\citep{niekum2012learning,konidaris2012robot,lioutikov2017learning,blank2024nils,blank2026sparc}.
These libraries are sequenced by planners or grammars
\citep{lioutikov2018pcfg,lioutikov2020attribute} rather than placed
after a chunk-predicting policy. The tolerance of a \proto{} plays the role
that the variance plays in a probabilistic movement primitive
\citep{paraschos2013probabilistic}, except that MPQ uses it as a
bound on execution rather than as a distribution to sample from. We
represent segments as B-spline control points following BEAST
\citep{zhou2025beast}, which tokenizes chunks without a learned encoder.
Other work places primitives \citep{scheikl2024mpd} or splines
\citep{tian2026splinepolicy,han2026bsplinepolicy}
inside the policy head. The
projection of cluster centroids onto real examples follows ProtoPNet
\citep{chen2019protopnet}. We introduce no new primitive. Our
contribution lies in where the library sits and in the measurement
this placement enables.

\paragraph{Execution-time constraints.}
Retrieval-based policies match \emph{observations} to demonstrations and
leave the executed action unconstrained
\citep{pari2021vinn,du2023behaviorretrieval,odonchimed2025ragdp}.
MPQ instead matches \emph{actions} at execution time, so that each
executed chunk has an identity and a deviation bound. Two further lines
of work act on the chunk at execution time. Bidirectional decoding
selects among chunks sampled from a policy \citep{liu2025bid}, and
residual policies learn corrections to a fixed controller or a
reference trajectory
\citep{silver2018residual,johannink2019residual,huang2025moreerl}. Like a residual
policy, MPQ corrects a fixed base, but the base is a demonstrated
segment and the correction is bounded rather than learned. Safety
shields \citep{alshiekh2018shielding} and control barrier functions
\citep{ames2019cbf} likewise project actions, but onto an implicit safe
set rather than an explicit library.

\section{\methodlong}
\label{sec:method}

\subsection{Measuring a vocabulary against a library \texorpdfstring{reference}{reference}}
\label{sec:instrument}

A vocabulary can be measured in closed loop only against a policy that
already succeeds without it. Let $\mathrm{SR}(\cdot)$ denote closed-loop
success rate, and let $\pi$ be a continuous reference policy that
succeeds on the benchmark. A \emph{library}
$\mathcal{P} = \{P_k\}_{k=1}^{K}$ is a finite set of executable chunks,
and an execution layer (\S\ref{sec:method:policy}) wraps $\pi$ through
this library. For each chunk that $\pi$ predicts, the layer selects a
library element, places it in the frame of the prediction and applies a
bounded correction.

The \emph{library reference} $\mathrm{SR}_{\mathrm{ref}}$ is the success
rate of $\pi$ executed through this layer. Since $\pi$ succeeds without
the layer, any drop in success can be attributed to the library.
$\mathrm{SR}_{\mathrm{ref}}$ is a baseline rather than an upper
bound, since a selector that chooses or corrects better could exceed
it. A discrete skill policy~$L$ over the same library replaces the
layer's decisions with learned heads. A choice head selects the
element, and further heads place and correct it. To isolate what its choice costs,
we define a hybrid $L_{\mathrm{idx}}$ that executes $L$'s choice with the
reference's placement and correction (\S\ref{sec:method:policy}). The
shortfall of $L$ then decomposes as
\begin{equation}
  \underbrace{\mathrm{SR}(\pi) - \mathrm{SR}(L)}_{\text{total gap}}
  \;=\;
  \underbrace{\mathrm{SR}(\pi) - \mathrm{SR}_{\mathrm{ref}}}_{\text{quantization}}
  \;+\;
  \underbrace{\mathrm{SR}_{\mathrm{ref}} - \mathrm{SR}(L_{\mathrm{idx}})}_{\text{selection}}
  \;+\;
  \underbrace{\mathrm{SR}(L_{\mathrm{idx}}) - \mathrm{SR}(L)}_{\text{execution}} .
  \label{eq:decomp}
\end{equation}
The quantization term is what the library cannot represent. The
selection term is what $L$'s choice costs when the reference supplies
the placement and correction, and the execution term is what $L$'s
learned placement and correction lose relative to the reference's.
Eq.~\ref{eq:decomp} substitutes $L$'s choice before its placement
and correction, and we call it the \emph{choice-first} order. In the
\emph{correction-first} order, a hybrid $L_{\mathrm{exec}}$ executes the
reference's choice with $L$'s placement and correction. \S\ref{sec:placement:results} reports both.

The decomposition holds
only if library elements execute without a learned decoder, since
otherwise part of any drop could come from the decoder. A latent
codebook does not meet this requirement
(Appendix~\ref{app:extra:codec}). The layer must also cost the
reference nothing, which we verify in \S\ref{sec:exp}.

\subsection{Representing action chunks}
\label{sec:method:spline}

We represent an action chunk $A \in \mathbb{R}^{H \times d}$ of horizon
$H$ over $d$ geometric degrees of freedom as a B-spline curve
\citep{zhou2025beast} with $B$ control points
$C \in \mathbb{R}^{B \times d}$. Evaluating the $B$ basis functions at
the $H$ timesteps gives the \emph{collocation matrix}
$\Phi \in \mathbb{R}^{H \times B}$. Decoding control points to a
trajectory is then the linear map $A \approx \Phi\, C$, and fitting
control points to a trajectory is its closed-form least-squares inverse.
Since neither operation is learned, every point of
$\mathbb{R}^{B \times d}$ decodes to a valid trajectory. The gripper is
fit on the same basis as a separate channel, with control points
$c^{\mathrm{g}} \in \mathbb{R}^{B}$.

Euclidean distance between control points does not equal the distance
between the trajectories they produce. The \emph{Gram matrix}
$M = \Phi^\top\!\Phi$ relates the two through
\begin{equation}
  \|\Phi(C - P)\|_F^2 \;=\; \operatorname{tr}\!\big((C-P)^\top M\, (C-P)\big)
  \;=:\; \|C - P\|_M^2 ,
  \label{eq:gram}
\end{equation}
so the trajectory distance can be computed on the coefficients without
decoding. We call
$\|\cdot\|_M$ the \emph{Gram metric} and compute it over geometric channels, comparing the gripper separately (Eq.~\ref{eq:invdist}).

\subsection{Canonicalizing orientation}
\label{sec:method:canon}

The same motion performed in different directions should match the same
library element. Because actions are end-effector deltas in a fixed base
frame, position is already factored out, and only the direction the
motion faces remains. We remove this direction with a
\emph{canonicalizing yaw rotation}~$\hat{R}_C$, which turns the net
horizontal displacement of chunk~$C$ to $+x$ and rotates position and
axis-angle deltas alike. For chunks with negligible horizontal
displacement, the rotation falls back to the principal horizontal axis
(Appendix~\ref{app:library}). Library elements are stored in this
canonical frame. Since the Gram metric acts along time, rotating a
trajectory leaves its norm unchanged, $\|\hat{R}\, C\|_M = \|C\|_M$. The
distance to a library element $P$ with gripper control points
$p^{\mathrm{g}}$ is
\begin{equation}
  d(C, P) \;=\; \big\| \hat{R}_C\, C - P \big\|_M^2
  \;+\; w_g^2\, \big\| c^{\mathrm{g}} - p^{\mathrm{g}} \big\|_M^2 ,
  \label{eq:invdist}
\end{equation}
where the weight $w_g$ separates an approach with the gripper opening
from a lift with it closing.

We factor out only yaw, because gravity distinguishes vertical from
horizontal motion. Under full 3D rotation, a vertical lift and a
horizontal push would map to the same skill. It also requires
rotation to act linearly on the actions, which holds for end-effector deltas but not in joint space.

\subsection{Building the library}
\label{sec:method:library}

\begin{figure}[t]
\centering
\begin{minipage}[b]{0.655\linewidth}
\centering
\includegraphics[width=\linewidth]{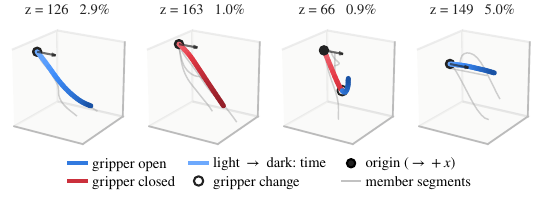}
\caption{Four \protos{} of the LIBERO-90 library, left to right a reach,
a closed-gripper transport, a gripper transition and a near-idle motion.
Headers: index $z$ and share of closed-loop usage by wrapped ACT. Full
gallery in Appendix~\ref{app:cards}.}
\label{fig:protos}
\end{minipage}\hfill
\begin{minipage}[b]{0.315\linewidth}
\centering
\includegraphics[width=\linewidth]{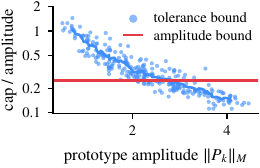}
\caption{Cap relative to amplitude, one point per LIBERO-90 \proto{}.
Blue line: running mean of the tolerance bound.}
\label{fig:bound}
\end{minipage}
\end{figure}

The library is extracted from demonstrations. Each demonstration is
split into overlapping $H$-step segments, which are canonicalized and
then clustered by $k$-means under the distance of Eq.~\ref{eq:invdist}.
Whitening by the Cholesky factor of $M$ turns the Gram metric into
Euclidean distance, so standard $k$-means applies. Each centroid is then
projected onto the nearest real segment, which makes every library
element a prototype of its cluster \citep{chen2019protopnet,li2024prototypes}. The
projection is needed because a cluster average can blend two valid
motions into one that is neither (Fig.~\ref{fig:protos}).

Each \proto{} also carries a \emph{tolerance}~$\tau_k$, defined as the
$q$-quantile ($q{=}0.75$) of the Gram distances from its cluster members
to~$P_k$. The tolerance captures how much the demonstrations vary for
that motion, and it is far from uniform. On LIBERO-90, small-amplitude
\protos{} have tolerances of $1.4\times$ their norm, whereas
large-amplitude ones have $0.35\times$ (Fig.~\ref{fig:bound}). Small,
precise motions such as aligning a gripper or correcting a grasp vary
more relative to their size than long transports across the workspace
do. An \emph{amplitude bound}, proportional to each element's norm,
therefore gives these motions the least room to correct, while our
\emph{tolerance bound}, set by $\tau_k$, gives them the most
(\S\ref{sec:exp:bound}). We choose $K$ by the library reference rather
than by reconstruction (\S\ref{sec:libraries:size}).

\subsection{Wrapping a policy}
\label{sec:method:policy}

MPQ attaches to any policy with a continuous chunk head. We write
this policy as $\pi$ and treat it as a frozen black box. Its predicted
chunk is fit to control points~$C_\pi$ and gripper control
points~$c^{\mathrm{g}}_\pi$. For each predicted chunk, the layer
\emph{selects} the nearest \proto{} $P_z$ under Eq.~\ref{eq:invdist},
\emph{places} it in the frame of the prediction as $R^{-1} P_z$ with
$R = \hat{R}_{C_\pi}$, and \emph{corrects} it toward the prediction by
an offset clipped to the \proto{}'s tolerance~$\tau_z$ scaled by a
budget~$\beta$. With $e = C_\pi - R^{-1}P_z$ denoting the gap between
the prediction and the placed \proto{}, the correction and the executed
chunk are
\begin{align}
  \Delta &= e \cdot \min\!\Big(1,\;\frac{\beta\, \tau_z}{\|e\|_M}\Big)
    & &\text{(clipped correction)}, \label{eq:clip} \\
  C^{\mathrm{exec}} &= R^{-1} P_z + \Delta
    & &\text{(executed chunk)}. \label{eq:bound}
\end{align}
The executed chunk therefore satisfies the \emph{deviation bound}
$\|C^{\mathrm{exec}} - R^{-1}P_z\|_M \le \beta\, \tau_z$. The amplitude
bound of \S\ref{sec:method:library} instead caps
$\|\Delta\|_M \le \beta \|P_z\|_M$. At $\beta{=}0$
the policy only chooses which element runs, and as $\beta \to \infty$
the layer reduces to the spline fit of the prediction. For $\beta{>}0$
the gripper passes through unchanged
($c^{\mathrm{g,exec}} = c^{\mathrm{g}}_\pi$),
so the policy decides when the gripper closes and the library where the
hand goes. Nothing on this path is learned.

The hybrid $L_{\mathrm{idx}}$ of Eq.~\ref{eq:decomp} replaces~$z$ by
another model's choice~$z_L$ and keeps the
reference's placement and correction,
\begin{equation}
  C^{\mathrm{exec}}_{L_{\mathrm{idx}}}
  \;=\; R_\pi^{-1}\, P_{z_L} + \Delta_\pi(z_L).
  \label{eq:lidx}
\end{equation}
This isolates the choice from placement and correction.

\section{Validating the execution layer}
\label{sec:exp}

The library reference measures a vocabulary only if the execution layer
costs a succeeding policy nothing. We test it across architectures,
benchmarks, observation shift and a real robot.

\subsection{Setup}
\label{sec:exp:setup}

Our main benchmark is LIBERO-90 \citep{liu2023libero}, whose 90 tasks
we evaluate at 10 episodes each from fixed initial states, with a cap of
600 steps per episode. LIBERO-10 and RoboCasa
\citep{nasiriany2024robocasa} are evaluated at 50 episodes per task
(Appendix~\ref{app:benchmarks}). Unless stated otherwise, the library
has $K{=}256$ \protos{} of $H{=}20$ steps, with $w_g{=}5$ and
$\beta{=}0.5$. These values were selected once for ACT on a disjoint
split of LIBERO-90 initial states (Appendix~\ref{app:noise}) and are
used unchanged for every policy, every suite and the real robot. An
\emph{in-suite} library is built from the benchmark's own
demonstrations, and a \emph{bare} policy runs without the layer at the
same replanning interval. Chunks longer than the library horizon are
passed to the layer as their first $H$ steps, which contain the executed
prefix (Appendix~\ref{app:extra:chunk}). Confidence intervals are 95\% bootstrap
intervals over tasks, and three-seed results are paired per seed.
Because repeated evaluation of one checkpoint varies by up to
3.5~points (Appendix~\ref{app:noise}), we do not interpret unpaired
differences below 3~points.

\subsection{The layer costs the reference nothing}
\label{sec:exp:main}

\begin{table}[t]
\caption{MPQ on six architectures. Success (\%) bare and through the in-suite library ($K{=}256$, $\beta{=}0.5$).
$H_\pi$: chunk length. Mean $\pm$ sd over three seeds, $\Delta$ paired
per seed. $\dagger$: released weights, one seed, $\Delta$ paired per
task with 95\% bootstrap CI. Pooled: 95\% $t$-interval over cells
(Appendix~\ref{app:pooled}).}
\label{tab:main}
\centering
\small
\setlength{\tabcolsep}{4pt}
\begin{tabular}{lccc@{\hskip 1em}r}
\toprule
policy & $H_\pi$ & bare & $+$MPQ & $\Delta$ \\
\midrule
\multicolumn{5}{l}{\emph{LIBERO-90}} \\
ACT & 20 & 78.8 {\tiny $\pm$ 2.3} & 80.2 {\tiny $\pm$ 1.0} & $+1.4$ {\tiny $\pm$ 2.1} \\
ACT, language-conditioned & 20 & 72.1 {\tiny $\pm$ 0.8} & 71.6 {\tiny $\pm$ 0.8} & $-0.6$ {\tiny $\pm$ 0.1} \\
Diffusion Policy & 16 & 91.7 {\tiny $\pm$ 0.6} & 92.9 {\tiny $\pm$ 1.2} & $+1.2$ {\tiny $\pm$ 1.8} \\
QueST & 32 & 85.0 {\tiny $\pm$ 0.7} & 82.9 {\tiny $\pm$ 2.1} & $-2.1$ {\tiny $\pm$ 1.5} \\
VQ-BeT & 5 & 69.6 {\tiny $\pm$ 6.1} & 68.3 {\tiny $\pm$ 5.1} & $-1.3$ {\tiny $\pm$ 1.2} \\
Flow-matching DiT & 16 & 81.9 {\tiny $\pm$ 1.5} & 83.4 {\tiny $\pm$ 1.4} & $+1.5$ {\tiny $\pm$ 1.6} \\
\midrule
\multicolumn{5}{l}{\emph{LIBERO-10}} \\
OpenVLA-OFT (7B)$^\dagger$ & 8 & 86.8 & 86.4 & $-0.4$ {\tiny[$-3.6$, $+2.6$]} \\
ACT & 20 & 64.7 {\tiny $\pm$ 3.4} & 70.3 {\tiny $\pm$ 1.9} & $+5.6$ {\tiny $\pm$ 2.9} \\
Diffusion Policy & 16 & 76.2 {\tiny $\pm$ 2.6} & 79.5 {\tiny $\pm$ 2.5} & $+3.3$ {\tiny $\pm$ 2.0} \\
Flow-matching DiT & 16 & 72.7 {\tiny $\pm$ 5.0} & 70.6 {\tiny $\pm$ 4.4} & $-2.1$ {\tiny $\pm$ 3.1} \\
\midrule
\multicolumn{4}{l}{pooled, 13 cells incl.\ RoboCasa (Table~\ref{tab:transfer})} & $+0.6$ {\tiny[$-0.8$, $+1.9$]} \\
\bottomrule
\end{tabular}
\end{table}

Table~\ref{tab:main} evaluates ACT \citep{zhao2023act}, Diffusion
Policy \citep{chi2023diffusion}, a flow-matching DiT
(Appendix~\ref{app:policies}), QueST \citep{mete2024quest}, VQ-BeT
\citep{lee2024vqbet}, and the 7B OpenVLA-OFT \citep{kim2025oft} with its
released LIBERO-10 weights. QueST and VQ-BeT are themselves discrete
skill policies, so for them the layer quantizes the output of their
decoders.

\paragraph{Cost of the layer.}
No policy loses more than evaluation noise. Since the spline fit
alone costs nothing (Appendix~\ref{app:extra:fitonly}), any cost comes
from quantization, even though the bound clips 67 to 93\% of chunks. The
largest losses are 1 to 2~points for QueST and VQ-BeT on LIBERO-90 and
$2.1 \pm 3.1$ for the flow DiT on LIBERO-10. Pooled over 13 cells with RoboCasa
(Table~\ref{tab:transfer}), success changes
by $+0.6$~points (95\% CI $[-0.8, +1.9]$). Two one-sided tests
show equivalence within $\pm 3$~points, the noise floor of
\S\ref{sec:exp:setup} ($p{<}0.001$, Appendix~\ref{app:pooled}). The layer also preserves smoothness
(Appendix~\ref{app:extra:smooth}).

\paragraph{Removing \protos{}.}
Removing the most-used \proto{} or five heavily used ones leaves
wrapped ACT within noise (seed~0, Appendix~\ref{app:extra:detector}).
Neighboring \protos{} substitute for each other within the correction,
so the library reference measures the whole library.

\paragraph{Imposing a constraint.}
Removing \protos{} can also impose a new constraint on a trained
policy without retraining it. As an example, we cap the commanded
speed, the largest translation a chunk commands per control step, at
the 75th percentile of the library and delete the 64 \protos{} above
it. Bare ACT exceeds the cap in 28.4\% of its chunks. With the
edited library, wrapped ACT exceeds it in 11.9\% at unchanged success
($79.3 \pm 1.2$ against $79.0 \pm 0.6$). Halving the bound limits how
much speed the correction can add and lowers violations to 4.5\%, at a
cost of 10~points (Appendix~\ref{app:edit:speed}).

\subsection{The deviation bound}
\label{sec:exp:bound}

\begin{figure}[t]
\centering
\includegraphics[width=0.92\linewidth]{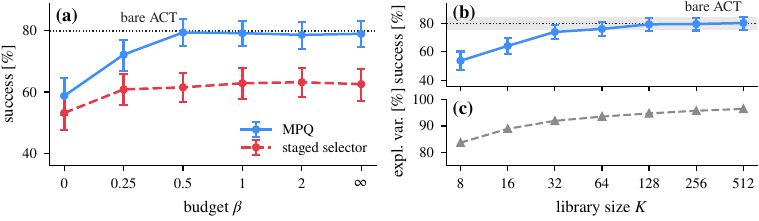}
\caption{Wrapped ACT on LIBERO-90 (seed~0). Error bars: 95\% bootstrap
CIs over tasks. \textbf{(a)} Success versus the budget $\beta$ for
MPQ and the staged selector (\S\ref{sec:analysis:staged}), both
with the $K{=}256$ library. Dotted: bare ACT. \textbf{(b)} Success
versus library size $K$ at $\beta{=}0.5$. Band: bare ACT with its 95\%
CI. \textbf{(c)} Offline explained variance of the same libraries.}
\label{fig:sweeps}
\end{figure}

Fig.~\ref{fig:sweeps}a varies $\beta$ for ACT (seed~0, bare 79.9). Success rises from 58.8 at $\beta{=}0$, where the policy only
selects the segment, to 79.4 at $\beta{=}0.5$, and stays flat for larger
budgets. Even at $\beta{=}0.5$, the tightest value on this plateau, the
bound clips 88\% of chunks. The median displacement from the placed
\proto{} is then 7\,cm, against 39\,cm commanded per one-second chunk
(Appendix~\ref{app:physical}).

The form of the bound matters as well. The relative spread of a \proto{}
falls with its amplitude, with a correlation of $-0.81$
(Fig.~\ref{fig:bound}), so the tolerance bound gives small motions more
room than an amplitude bound does. Compared with the amplitude bound
and a gripper scaled by the clip, the tolerance bound with the gripper
passed through raises the paired difference to
the bare policy on each of the nine three-seed LIBERO rows of
Table~\ref{tab:main}, by 3.0~points on average
(Appendix~\ref{app:bound:rows}).

\subsection{Execution on a real robot}
\label{sec:exp:robot}
\begin{wrapfigure}[15]{r}{0.28\linewidth}
\vspace{-\intextsep}
\centering
\includegraphics[width=\linewidth]{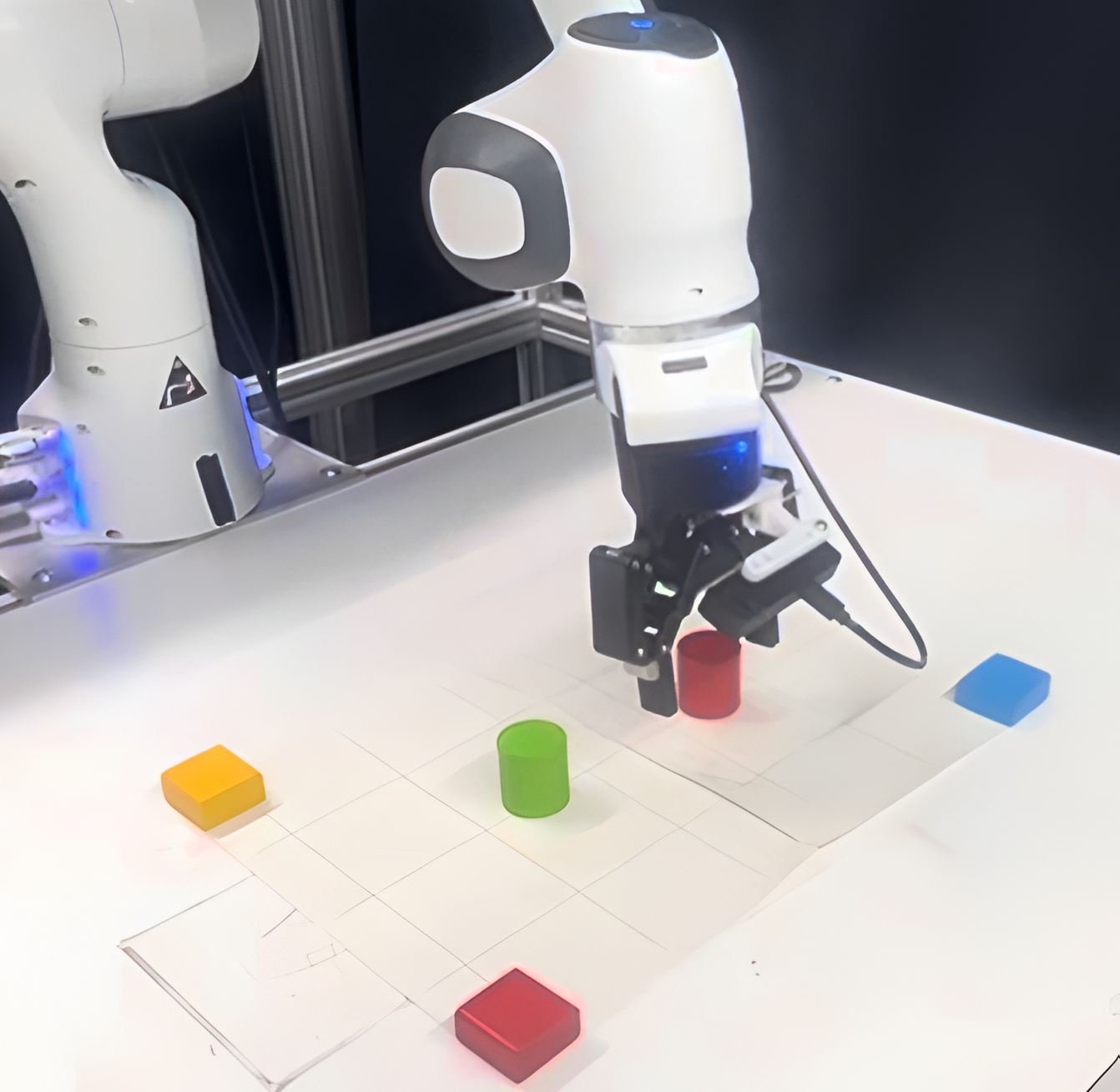}
\caption{Real-robot task. Stacking a cylinder on the instructed
cube.}
\label{fig:robot}
\end{wrapfigure}
We run a fine-tuned $\pi_{0.5}$ \citep{intelligence2025pi05} on a
Franka Panda in the DROID configuration \citep{khazatsky2024droid}
(Appendix~\ref{app:robot}). The task is to stack one of two cylinders on
whichever of four colored cubes the instruction names. The library is
built from the 400 fine-tuning demonstrations with the defaults of
\S\ref{sec:exp:setup}, and the bound is the one selected in simulation.
Since the policy commands absolute poses, the layer converts them to
deltas and back. The bare policy succeeds in 15 of 15 attempts and the
wrapped policy in 13 of 15 (Fisher exact test, $p{=}0.48$). Both
failures are missed grasps in which the gripper stops above the
cylinder. With fifteen attempts per arm, this comparison can resolve
only large differences. Offline, on held-out demonstrations of the same
task, the tolerance bound executes 81\% of a 1\,cm lateral correction
before the release, compared with 63\% for the amplitude bound
(Appendix~\ref{app:robot}).

\subsection{Robustness under observation shift}
\label{sec:exp:shift}

\begin{table}[t]
\caption{Observation shift on LIBERO-Plus, LIBERO-10 policies
($K{=}256$ in-suite library, $\beta{=}0.5$). In distribution: ten unperturbed
tasks, 10 initial states each (Table~\ref{tab:main}: 50). Shifted: 6 axes
$\times$ 100 variants. Mean
$\pm$ sd over three seeds, $\Delta$ paired per seed. Per axis in
Appendix~\ref{app:extra:plus}.}
\label{tab:plus}
\centering
\small
\setlength{\tabcolsep}{4pt}
\begin{tabular}{lccrccr}
\toprule
 & \multicolumn{3}{c}{in distribution} & \multicolumn{3}{c}{six shifted axes} \\
\cmidrule(lr){2-4} \cmidrule(lr){5-7}
policy & bare & $+$MPQ & $\Delta$ & bare & $+$MPQ & $\Delta$ \\
\midrule
Diffusion Policy & 85.8 {\tiny $\pm$ 1.0} & 84.3 {\tiny $\pm$ 1.5} & $-1.5$ {\tiny $\pm$ 2.3} & 36.6 {\tiny $\pm$ 1.4} & 37.9 {\tiny $\pm$ 1.5} & $+1.3$ {\tiny $\pm$ 2.4} \\
ACT & 67.7 {\tiny $\pm$ 9.0} & 77.0 {\tiny $\pm$ 6.6} & $+9.3$ {\tiny $\pm$ 6.8} & 29.5 {\tiny $\pm$ 1.8} & 33.5 {\tiny $\pm$ 1.6} & $+4.0$ {\tiny $\pm$ 2.1} \\
Flow-matching DiT & 81.0 {\tiny $\pm$ 3.6} & 80.7 {\tiny $\pm$ 3.8} & $-0.3$ {\tiny $\pm$ 3.2} & 33.7 {\tiny $\pm$ 2.1} & 33.1 {\tiny $\pm$ 4.4} & $-0.6$ {\tiny $\pm$ 3.4} \\
\bottomrule
\end{tabular}
\end{table}

LIBERO-Plus \citep{fei2025liberoplus} perturbs the LIBERO-10 tasks
along six axes, namely viewpoint, initial state, noise, layout, texture
and lighting. We evaluate the three LIBERO-10 policies, which were
trained without augmentation, on 100 paired variants per axis.

Under shift, the layer costs no more than in distribution
(Table~\ref{tab:plus}). On no single axis does any policy show a paired
loss whose interval excludes zero (Appendix~\ref{app:extra:plus}). This holds although shift moves bare ACT's
chunks farther from the demonstrations on every axis, even in successful
episodes
(Appendix~\ref{app:physical}). The layer keeps each chunk within the bound of
a demonstrated one.

\section{Decomposing a discrete selector's gap}
\label{sec:placement}

Since the execution layer costs a succeeding policy nothing
(\S\ref{sec:exp}), the gap between a continuous reference and a discrete
selector decomposes into quantization, selection and execution as in
Eq.~\ref{eq:decomp}. We decompose selectors on three references, ACT,
the flow DiT and Diffusion Policy.

\subsection{\texorpdfstring{Selectors}{Selectors}}
\label{sec:analysis:router}
Every selector $L$ replaces a policy's continuous action output by
the skill heads of \S\ref{sec:instrument}. The heads are trained on
demonstrations labeled with their nearest \proto{}, after the library
is built and frozen (Appendix~\ref{app:selectors}). Each selector
keeps the trained image and state encoder of its reference frozen and
trains only a readout and the skill heads (Table~\ref{tab:decomp}).
Selector and reference thus see the same features, and the
decomposition isolates the cost of the discrete output. The heads
generate placement and correction as the reference generates actions.
ACT's selector regresses them, the flow DiT's samples them by flow
matching and Diffusion Policy's by diffusion.

\begin{table}[t]
\caption{Decomposition of the selectors' gaps (LIBERO-90,
$K{=}256$, $\beta{=}0.5$). Heads: how the selector generates placement
and correction. $\mathrm{SR}(\pi)$ and quantization as in
Table~\ref{tab:main}. Mean $\pm$ sd over three seeds.}
\label{tab:decomp}
\centering
\small
\setlength{\tabcolsep}{3.2pt}
\begin{tabular}{llcccrrrr}
\toprule
 & & & & & \multicolumn{2}{c}{choice first} & \multicolumn{2}{c}{correction first} \\
\cmidrule(lr){6-7} \cmidrule(lr){8-9}
reference & heads & $\mathrm{SR}(\pi)$ & $\mathrm{SR}(L)$ & quant. & selection & execution & execution & selection \\
\midrule
ACT & regression & 78.8 & 67.1 {\tiny $\pm$ 3.5} & $+1.4$ & $-0.9$ {\tiny $\pm$ 1.2} & $\mathbf{-12.1}$ {\tiny $\pm$ 1.6} & $\mathbf{-12.4}$ {\tiny $\pm$ 2.1} & $-0.6$ {\tiny $\pm$ 2.1} \\
flow DiT & flow & 81.9 & 83.0 {\tiny $\pm$ 0.0} & $+1.5$ & $-0.4$ {\tiny $\pm$ 1.0} & $\mathbf{+0.0}$ {\tiny $\pm$ 2.0} & $\mathbf{-3.1}$ {\tiny $\pm$ 1.1} & $+2.7$ {\tiny $\pm$ 1.2} \\
Diff.\ Policy & diffusion & 91.7 & 82.0 {\tiny $\pm$ 2.0} & $+1.2$ & $-0.5$ {\tiny $\pm$ 1.1} & $\mathbf{-10.3}$ {\tiny $\pm$ 1.9} & $\mathbf{-9.0}$ {\tiny $\pm$ 1.5} & $-1.8$ {\tiny $\pm$ 2.5} \\
\bottomrule
\end{tabular}
\end{table}

\subsection{\texorpdfstring{Selection and execution terms}{Selection and execution terms}}
\label{sec:exp:decomp}
\label{sec:placement:results}
\begin{wrapfigure}[16]{r}{0.46\linewidth}
\vspace{-\intextsep}
\centering
\includegraphics[width=\linewidth]{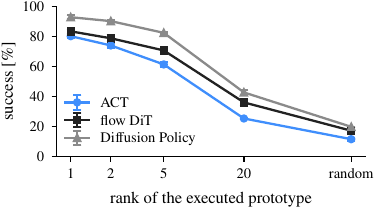}
\vspace{-1.6em}
\caption{Success when the reference executes its $r$-th nearest
\proto{} (random: expected rank 128). LIBERO-90, $K{=}256$,
$\beta{=}0.5$, three seeds.}
\label{fig:selection}
\end{wrapfigure}
\leavevmode{}Table~\ref{tab:decomp} shows that the selectors' gaps arise in executing the chosen
element, not in choosing it. Quantization is
within noise for all three references with this library, although it
grows for smaller ones (\S\ref{sec:libraries:size}). In the choice-first order,
selection changes success by at most 0.9~points. Execution costs the
selectors of ACT and Diffusion Policy 12.1 and 10.3~points, nearly
their entire gaps, whereas the flow DiT's selector loses nothing in
execution and reaches its library reference (83.0 against 83.4). The
correction-first order moves at most 3.1~points between the two
terms. Appendix~\ref{app:selectors} gives bootstrap intervals for
all terms.\\[\parskip]
A small selection term could also mean that the reference's
correction compensates for any choice. We therefore force the layer to
execute the $r$-th nearest \proto{} to the reference's chunk, or a random
one, under the reference's placement and correction
(Fig.~\ref{fig:selection}). Success falls
steeply with rank. For ACT, the fifth-nearest \proto{} costs
$18.9 \pm 2.6$~points and a random one $68.7 \pm 1.7$. The flow DiT and
Diffusion Policy follow the same curve.
Without any correction ($\beta{=}0$, Appendix~\ref{app:selectors}), the fifth-nearest \proto{} costs
about 25~points, while the choices of the staged ACT selector
(\S\ref{sec:analysis:staged}) cost 2.9. The
selection term is therefore small because the selectors choose well,
not because the correction hides their choices.

\subsection{\texorpdfstring{Source of the execution loss}{Source of the execution loss}}
\label{sec:analysis}
\label{sec:analysis:staged}
ACT's execution loss follows from how placement and correction are computed.
MPQ derives both from the reference's continuous prediction,
so the executed chunk follows that prediction as closely as the bound
allows, whichever element was chosen. ACT's selector instead
regresses yaw and correction from the observation alone, with a single
offset that does not depend on the chosen element. The analyses
below use the \emph{staged} selector, ACT's network with these heads
trained end to end. It loses 14.6~points in execution, against 12.1 for
the frozen selector of Table~\ref{tab:decomp}, so training the encoder
on the selection objective changes the loss little. On held-out
chunks, its predicted yaw has a median error of $11^\circ$
(Appendix~\ref{app:selectors}).

\paragraph{Element-conditioned heads.}
The staged selector's placement and correction heads do not see the chosen
\proto{}, which could by itself explain the execution loss. We therefore
condition these heads on an embedding of the chosen
\proto{} and compare them with heads of equal capacity without it
(Appendix~\ref{app:selectors}). At execution the heads see the selector's own
choice. Conditioning raises success from 65.6 to about 70, whether the
heads are trained on the labeled \proto{} or on the selector's own
choices. The selector still falls about 10~points short of the library
reference, and execution accounts for 9 to 10 of them. The element-independent heads therefore explain only a small
part of the execution loss. Diffusion Policy's selector, whose heads
see the chosen element, still loses 10.3~points in execution.

\paragraph{Controls.}
Removing the bound does not help. Without
it, the staged selector plateaus at 61 to 63 while MPQ
reaches 79 (Fig.~\ref{fig:sweeps}a), so its loss does not come from
the bound. A correction network of Diffusion Policy's own width raises its
selector from 82.0 to 85.8, a third of its gap, but execution still
costs 6.3~points (Appendix~\ref{app:selectors}).
Selectors that co-adapt the \protos{} under a curriculum
\citep{li2023curriculum,blessing2023imc} or execute a sparse mixture of
them, as mixture-of-experts policies do
\citep{zhou2024vdd,reuss2025mode}, score 21 to 33~points below the
staged selector (Appendix~\ref{app:selectors}). QueST scores 85.0, above
bare ACT, but cannot be decomposed, because its codes need a learned
decoder (Appendix~\ref{app:extra:codec}).

\section{Closed-loop versus offline library evaluation}
\label{sec:libraries}
\label{sec:analysis:offline}

Libraries are typically validated offline, by the share of
demonstration-action variance they reconstruct
\citep{zhou2025beast,lee2024vqbet,mete2024quest}. The library reference instead
evaluates them in closed loop. We compare the two criteria on library size,
library design and transfer.

\subsection{Library size and design}
\label{sec:libraries:size}
\label{sec:exp:ablations}

Wrapped ACT matches bare ACT from $K{=}128$ onward, while
reconstruction keeps improving by another 1.7~points over two further
doublings (Fig.~\ref{fig:sweeps}b and~\ref{fig:sweeps}c). Selecting $K$ by reconstruction would
therefore choose a library several times larger than the task requires.

The two criteria also disagree on which design choices matter
(Table~\ref{tab:ablations}). Dividing out amplitude in addition to yaw
costs 5~points in closed loop, yet reconstruction ranks
the resulting library highest. Yaw canonicalization itself adds about
3~points of success at $K{=}256$ and 10 at $K{=}64$
(Appendix~\ref{app:design}). The Gram metric, the clustering algorithm
and the projection onto real segments each change success by less than
the noise at $K{=}256$, and our $k$-means library comes within 3~points
of executing through all 118{,}525 segments. Leaving out that last row,
reconstruction varies by 3.1~points across the libraries of
Table~\ref{tab:ablations}, while closed-loop success varies by 5.5.

\begin{table}[t]
\centering
\footnotesize
\setlength{\tabcolsep}{2.5pt}
\begin{minipage}[t]{0.455\linewidth}
\caption{Library ablations on LIBERO-90. ACT success and offline
explained variance (\%) ($K{=}256$, $w_g{=}5$,
$\beta{=}0.5$, seed~0, bare 79.9). Brackets: 95\% bootstrap CIs over
tasks. Ours: $k$-means, projected.}
\label{tab:ablations}
\centering
\begin{tabular}{lcc}
\toprule
library & success & offline \\
\midrule
Euclidean & 74.6 {\tiny[69.7, 79.1]} & 93.6 \\
Gram metric (Eq.~\ref{eq:gram}) & 76.7 {\tiny[71.9, 81.1]} & 94.0 \\
\quad $+$ yaw (ours) & 79.4 {\tiny[74.9, 83.8]} & 95.6 \\
\quad $+$ amplitude & 74.4 {\tiny[69.2, 79.4]} & 95.9 \\
\midrule
EM-GMM, projected & 77.6 {\tiny[72.8, 82.0]} & 95.1 \\
256 random segments & 73.9 {\tiny[68.9, 78.8]} & 92.8 \\
all 118{,}525 segments & 82.1 {\tiny[78.0, 85.9]} & 100.0 \\
\bottomrule
\end{tabular}
\end{minipage}\hfill
\begin{minipage}[t]{0.535\linewidth}
\caption{Library transfer. Bare success and its change through the
target suite's own library and through the LIBERO-90 library
($K{=}256$, $\beta{=}0.5$). Mean $\pm$ sd over three seeds, $\Delta$
paired per seed.}
\label{tab:transfer}
\centering
\begin{tabular}{lccc}
\toprule
 & & \multicolumn{2}{c}{$\Delta$ through library} \\
\cmidrule(lr){3-4}
policy & bare & own & LIBERO-90 \\
\midrule
\multicolumn{4}{l}{\emph{RoboCasa}} \\
ACT & 43.6 {\tiny $\pm$ 4.2} & $-0.9$ {\tiny $\pm$ 2.4} & $-7.9$ {\tiny $\pm$ 1.2} \\
Diffusion Policy & 43.4 {\tiny $\pm$ 0.3} & $-0.1$ {\tiny $\pm$ 3.9} & $-4.2$ {\tiny $\pm$ 1.8} \\
Flow-matching DiT & 39.8 {\tiny $\pm$ 1.3} & $+1.7$ {\tiny $\pm$ 2.3} & $-6.5$ {\tiny $\pm$ 1.5} \\
\midrule
\multicolumn{4}{l}{\emph{LIBERO-10}} \\
ACT & 64.7 {\tiny $\pm$ 3.4} & $+5.6$ {\tiny $\pm$ 2.9} & $-0.2$ {\tiny $\pm$ 0.6} \\
\bottomrule
\end{tabular}
\end{minipage}
\end{table}

\subsection{Transfer across suites}
\label{sec:libraries:transfer}
\label{sec:exp:robocasa}

Transfer separates the two criteria most clearly
(Table~\ref{tab:transfer}). On RoboCasa, the LIBERO-90 library costs the
three policies 4 to 8~points, whereas the in-suite library costs nothing
within noise. Reconstruction nevertheless rates them
within 3~points (89.0\% against 92.0\%). On LIBERO-10, the
LIBERO-90 library executes ACT at parity with the bare policy
($-0.2 \pm 0.6$) but misses the 5.6-point gain that LIBERO-10's own
library provides. Reconstruction again rates the
two within 0.1\%.

Coverage distinguishes the two cases. The RoboCasa demonstrations
effectively occupy only 100 of the 256 \protos{} in the LIBERO-90
library, against 167 of their own.
The LIBERO-10 demonstrations occupy 165 and 186
(Appendix~\ref{app:extra:offline}). The RoboCasa loss is therefore a
coverage failure. The LIBERO-10 loss is not, and neither reconstruction
nor coverage detects it.

A coverage failure can be repaired without touching the policy.
Re-clustering both corpora together at $K{=}256$ brings RoboCasa within
1.2~points of its own library (seed~0, $[-5.5, +2.7]$) and leaves
LIBERO-90 unchanged ($-0.4 \pm 0.6$, three seeds). Appending RoboCasa
\protos{} to the LIBERO-90 library recovers less
(Appendix~\ref{app:extra:offline}).

The two criteria thus answer different questions. Reconstruction asks whether a library
can represent the demonstrations, the library reference whether the element
executed in a state is the right one.

\section{Conclusion}
\label{sec:conclusion}

A discrete skill policy combines a vocabulary of motions, a learned
choice among its elements and learned heads that execute it.
Measured separately in closed
loop, their costs differ widely. For selectors on three backbones, the
vocabulary costs little, and so does the choice when the reference
places and corrects it. Most of the loss lies in the
learned placement and correction. Conditioning a regression head on the chosen element
recovers little of this loss. When a discrete, editable library is required, wrapping a trained
continuous policy is therefore at least as effective as learning a
discrete head.

Placing the vocabulary after the policy has further consequences. Every executed chunk stays within a bound of a demonstrated
motion, and the library can be built, evaluated and edited while the
policy stays frozen. Offline reconstruction cannot guide such edits,
because it does not resolve closed-loop differences across library
sizes, designs or domains. The library reference resolves them at the cost
of one evaluation run per design. For a vision-language-action model that emits
an index over such a library, the execution term isolates what the
model must supply beyond the index.

\paragraph{Limitations.}
Our experiments use one arm type and a single real-robot task. The
decomposition applies to other selector architectures, but we tested
only selectors built on a reference's network or features. Finally,
the library reference is relative to a reference policy rather than an
optimal one, and the bound constrains the executed motion but not its effect on
the scene.

\subsection*{Acknowledgments}
The work was funded by the German Research Foundation (DFG) -- 448648559. The authors gratefully acknowledge the support of the Robotics Institute Germany (RIG).

\subsection*{AI use statement}
In this work, we used generative AI tools (Anthropic's Claude models) to
implement methods and to provide feedback on experiments. We have not used
generative AI tools to generate synthetic data sets, to develop theoretical
models or conceptual frameworks, to formulate mathematical claims, to
propose or refine hypotheses, to clean or reformat data sets, or to
interpret results. Proofs, translation, and qualitative or thematic
data analysis are not applicable to this work.

Additionally, we used generative AI tools to create and debug software code, including figure
plotting code, to search for and identify relevant literature, to format
references, to draft parts of the research paper and to edit the paper for readability.

We have reviewed all AI-assisted work. In particular, the authors reviewed generated code and
unit tests, traced every number reported in the paper to a logged
evaluation run, and checked every citation against its published version.

We take responsibility for the final content of this work, including
text, claims or artifacts produced with the aid of generative AI.

\subsection*{Reproducibility statement}
All code for library construction, execution and
evaluation is available at \url{https://intuitive-robots.github.io/mpq}.
Appendix~\ref{app:repro} details the fixed evaluation
protocols for all three benchmarks, the seed and noise-floor conventions,
and all hyperparameters.

All experiments ran on desktop workstations with consumer GPUs
(NVIDIA RTX 3070, 4080, and 5090) and CPUs (Intel Core i7 and
AMD Ryzen 9). The project used roughly 500 GPU hours in total,
including exploratory runs. Closed-loop evaluation is CPU-bound and uses
the GPU only for policy inference.

\bibliography{iclr2027_conference}
\bibliographystyle{iclr2027_conference}

\clearpage
\appendix

\section{Experimental details}
\label{app:repro}

\subsection{Chunks, spline fit and library construction}
\label{app:spline}
\label{app:library}
Demonstrations are cut into windows of $H{=}20$ actions at stride 5,
which gives 118{,}525 segments on LIBERO-90. Each window is fit to
$B{=}10$ B-spline control points with BEAST's knot configuration
\citep{zhou2025beast}, all channels on the same basis. The fit is
near-lossless (residual MSE $1.29\times10^{-3}$, per-step action variance
$2.01\times10^{-1}$).

Canonicalization rotates each segment about the gravity axis until its
net horizontal displacement points along $+x$. For displacements shorter
than $10^{-3}$ action units, the first principal axis of the per-step
horizontal displacements is aligned instead, and if these also vanish
the yaw is zero. The gripper channel is never rotated. The same rule
canonicalizes the policy's chunks at execution time. $k$-means runs on
Cholesky-whitened coefficients, with the whitened gripper channel scaled
by $w_g$, and each centroid is projected onto its nearest segment.
Building a library takes about two minutes on a desktop workstation.

\subsection{Policies}
\label{app:policies}
\label{app:training}
ACT, Diffusion Policy and the flow-matching policy are implemented in
LeRobot \citep{cadene2024lerobot}.

\textbf{ACT} and all selectors share one training recipe. They use an
ImageNet-pretrained ResNet-18 with frozen BatchNorm and an ACT-style
transformer \citep{zhao2023act}, and they train for 50k steps at batch
size 128 with AdamW (learning rate and weight decay $10^{-4}$) and a
cosine schedule. Task-id conditioning indexes an embedding table, and
language conditioning projects frozen CLIP ViT-B/16 sentence embeddings
of the instruction.

\textbf{Diffusion Policy} \citep{chi2023diffusion} uses a ResNet-18 per
camera and a 1D-UNet denoiser over 16-step chunks with one observation
step, with the task one-hot appended to proprioception. It is trained
for 100k steps at batch size 64 (AdamW, lr $10^{-4}$, 500-step warmup,
cosine decay) with $112 \times 112$ crops and DDPM with 100 noise steps.
We evaluate it with the EMA weights (decay 0.9995) and 10 DDIM steps.
Proprioception enters unnormalized, which affects absolute success but
not the paired differences.

The \textbf{flow-matching policy} is an 8-layer diffusion transformer
(hidden size 768, 12 heads, conditioning MLP of width 512) over an
ImageNet-pretrained ResNet-18. It predicts 16-step chunks and is trained
for 100k steps at batch size 64 without crops.

\textbf{QueST and VQ-BeT} are trained through their official pipelines
at the default configuration. For VQ-BeT, this means 5-step chunks, with
the codebook and the head trained jointly.

\textbf{OpenVLA-OFT} is the released LIBERO-10 checkpoint, served
through the authors' inference server with their settings (L1-regression
head, proprioception, center crop, agentview and wrist images, 8-step
chunks, open loop), evaluated at a single seed. It scores 86.8 against
the reported 94.5. Our protocol differs in settling steps (5 vs.\ 10)
and step cap (600 vs.\ 520), and the bare and wrapped cells share it.

\subsection{Benchmarks and evaluation protocol}
\label{app:benchmarks}

\textbf{LIBERO-90} has 90 tasks, each evaluated for 10 episodes from the
fixed initial states (900 rollouts per cell) with a 600-step cap.
\textbf{LIBERO-10} (LIBERO-Long) has 10 long-horizon tasks, evaluated at
50 initial states each (500 rollouts). For \textbf{RoboCasa} we use the
17 arm-driven atomic tasks \citep{nasiriany2024robocasa} with the mobile
base idle, at 50 episodes per task (850 rollouts). The RoboCasa policies
and libraries use all 500 machine-generated demonstrations per task.
Both simulators drive a Franka Panda through an operational-space
controller, so transfer between them crosses domains but not
embodiments.

\textbf{Replanning and latency.} ACT and OpenVLA-OFT execute their full
chunks of 20 and 8 steps. The Diffusion Policy and the flow-matching
policy execute 8 of 16 steps, QueST 8 of 32 and VQ-BeT 2 of 5, both bare
and wrapped. The layer quantizes the full predicted chunk before the
executed prefix runs. This adds 0.3\,ms per chunk on one CPU thread. The
overhead is flat in $K$ from 64 to 512, is dominated by the spline fit
and amounts to about 0.03\% of a chunk's execution time at 20\,Hz.

\subsection{Seeds, noise floor and selection split}
\label{app:noise}
Three evaluations of wrapped ACT (seed 0) through the default library
gave 80.0, 79.4 and 79.0. Two evaluations of the Euclidean cell of
Table~\ref{tab:ablations} gave 71.1 and 74.6, the largest spread we
observed. We therefore treat differences under 3 points as noise.

The design sweeps (Tables~\ref{tab:ablations} and
\ref{app:tab:designs}, the $\beta$ and $K$ sweeps, and
\S\ref{sec:analysis}) evaluate seed-0 checkpoints against the seed-0
reference of 79.9. The headline results (Tables~\ref{tab:main},
\ref{tab:decomp} and \ref{tab:transfer}) average three seeds, with the
library reference and the selectors evaluated per seed.
Table~\ref{app:tab:main} gives the per-seed values behind
Table~\ref{tab:main}.

\textbf{Selection split.} Every LIBERO-90 cell uses the first 10 of the
50 initial states per task, and hyperparameters are selected on states
10 to 19. On these states, wrapped ACT (seed 0) scores 72.4 / 80.4 /
79.4 for $\beta \in \{0.25, 0.5, 1\}$ at $q{=}0.75$ and 74.6 / 80.3 /
76.6 at $q{=}0.9$ (task-level intervals about $\pm 4.5$), so
$\beta{=}0.5$ is best at either quantile. $K$ and $w_g$ were selected
on the same split under the amplitude bound, where all cells with
$K \ge 256$ lie within 3 points, and kept for the tolerance bound.

\begin{table}[h]
\caption{Per-seed values behind Table~\ref{tab:main}, including its
RoboCasa rows ($K{=}256$, $\beta{=}0.5$, $q{=}0.75$). Brackets: 95\%
bootstrap CIs over tasks. $^\ast$: ablation of
Appendix~\ref{app:extra:chunk}, not part of Table~\ref{tab:main}.}
\label{app:tab:main}
\centering
\small
\begin{tabular}{llcc}
\toprule
policy & seed & bare & $+$MPQ \\
\midrule
\multicolumn{4}{l}{\emph{LIBERO-90}} \\
ACT, task-id & 0 & 79.9 {\tiny[75.2, 84.2]} & 79.4 {\tiny[74.9, 83.8]} \\
 & 1 & 80.3 {\tiny[75.9, 84.4]} & 81.3 {\tiny[77.3, 85.0]} \\
 & 2 & 76.1 {\tiny[70.9, 81.1]} & 79.8 {\tiny[75.1, 84.1]} \\
ACT, language & 0 & 71.3 {\tiny[66.1, 76.3]} & 70.7 {\tiny[64.8, 76.2]} \\
 & 1 & 72.9 {\tiny[67.7, 77.9]} & 72.3 {\tiny[66.8, 77.7]} \\
 & 2 & 72.2 {\tiny[66.8, 77.2]} & 71.7 {\tiny[66.0, 77.1]} \\
Diffusion Policy & 0 & 92.2 {\tiny[90.1, 94.1]} & 92.1 {\tiny[89.7, 94.3]} \\
 & 1 & 91.8 {\tiny[89.4, 93.9]} & 92.2 {\tiny[89.7, 94.6]} \\
 & 2 & 91.0 {\tiny[88.8, 93.0]} & 94.2 {\tiny[92.0, 96.2]} \\
Flow-matching DiT & 0 & 83.3 {\tiny[79.3, 87.0]} & 84.8 {\tiny[80.8, 88.4]} \\
 & 1 & 80.3 {\tiny[76.2, 84.2]} & 83.4 {\tiny[79.3, 87.1]} \\
 & 2 & 82.0 {\tiny[77.9, 85.8]} & 82.0 {\tiny[77.7, 85.9]} \\
QueST & 0 & 84.8 {\tiny[80.2, 89.0]} & 81.2 {\tiny[76.0, 86.0]} \\
 & 1 & 85.8 {\tiny[82.2, 89.0]} & 85.2 {\tiny[81.4, 88.8]} \\
 & 2 & 84.4 {\tiny[80.7, 87.9]} & 82.3 {\tiny[77.8, 86.6]} \\
QueST, full 32-step chunks$^\ast$ & 0 & 83.0 {\tiny[78.4, 87.1]} & 69.8 {\tiny[63.4, 75.7]} \\
 & 1 & 87.2 {\tiny[84.3, 89.9]} & 72.9 {\tiny[66.9, 78.6]} \\
 & 2 & 85.6 {\tiny[81.8, 89.1]} & 72.6 {\tiny[66.4, 78.3]} \\
VQ-BeT & 0 & 63.1 {\tiny[56.6, 69.6]} & 63.1 {\tiny[56.9, 69.2]} \\
 & 1 & 75.1 {\tiny[70.6, 79.4]} & 73.3 {\tiny[68.7, 77.8]} \\
 & 2 & 70.6 {\tiny[65.0, 76.0]} & 68.3 {\tiny[62.2, 74.2]} \\
\midrule
\multicolumn{4}{l}{\emph{LIBERO-10}} \\
OpenVLA-OFT & 0 & 86.8 {\tiny[79.4, 93.2]} & 86.4 {\tiny[79.4, 92.6]} \\
ACT, task-id & 0 & 63.6 {\tiny[54.2, 74.0]} & 71.8 {\tiny[63.4, 80.2]} \\
 & 1 & 68.6 {\tiny[60.4, 78.6]} & 71.0 {\tiny[62.8, 79.2]} \\
 & 2 & 62.0 {\tiny[51.0, 74.0]} & 68.2 {\tiny[60.6, 76.0]} \\
Diffusion Policy & 0 & 78.8 {\tiny[68.2, 88.6]} & 82.2 {\tiny[73.2, 90.4]} \\
 & 1 & 73.6 {\tiny[62.4, 84.8]} & 78.8 {\tiny[69.0, 88.2]} \\
 & 2 & 76.2 {\tiny[66.6, 85.0]} & 77.4 {\tiny[67.4, 87.2]} \\
Flow-matching DiT & 0 & 74.2 {\tiny[61.8, 86.0]} & 68.6 {\tiny[55.0, 82.2]} \\
 & 1 & 67.2 {\tiny[53.8, 79.8]} & 67.6 {\tiny[55.2, 80.2]} \\
 & 2 & 76.8 {\tiny[66.0, 87.2]} & 75.6 {\tiny[65.8, 84.8]} \\
\midrule
\multicolumn{4}{l}{\emph{RoboCasa} (in-suite library, $K{=}256$)} \\
ACT & 0 & 48.1 {\tiny[37.4, 59.4]} & 44.7 {\tiny[34.5, 55.5]} \\
 & 1 & 42.9 {\tiny[31.9, 54.7]} & 42.4 {\tiny[31.9, 53.5]} \\
 & 2 & 39.8 {\tiny[28.7, 51.6]} & 41.2 {\tiny[30.0, 53.3]} \\
Diffusion Policy & 0 & 43.5 {\tiny[31.3, 56.4]} & 42.1 {\tiny[31.5, 53.6]} \\
 & 1 & 43.1 {\tiny[31.6, 55.1]} & 39.9 {\tiny[27.9, 52.9]} \\
 & 2 & 43.6 {\tiny[32.7, 55.4]} & 47.9 {\tiny[36.1, 60.1]} \\
Flow-matching DiT & 0 & 40.5 {\tiny[29.4, 52.2]} & 40.1 {\tiny[29.5, 51.4]} \\
 & 1 & 38.4 {\tiny[28.2, 49.4]} & 42.6 {\tiny[31.8, 53.8]} \\
 & 2 & 40.6 {\tiny[30.1, 51.5]} & 41.9 {\tiny[31.2, 52.7]} \\
\bottomrule
\end{tabular}
\end{table}

\section{Amplitude and tolerance bounds per policy}
\label{app:bound:rows}
Table~\ref{app:tab:boundrows} compares the layer of Table~\ref{tab:main}
with an amplitude-bound configuration on every three-seed LIBERO row.
That configuration caps the correction at $0.25\,\|P_z\|_M$ and
scales the gripper channel with the same clip factor, whereas the
layer of Table~\ref{tab:main} uses the tolerance bound and passes the gripper through
(\S\ref{sec:method:policy}). At about equal tightness on ACT (seed~0),
the tolerance bound at $q{=}0.9$ clips 74\% of chunks and scores 80.6,
the amplitude bound clips 78\% and scores 78.0, and a constant cap
(95\% clipped) scores 78.1. These single cells differ by less than the
noise floor, but the direction holds across the policies of
Table~\ref{app:tab:boundrows}.

\begin{table}[h]
\caption{Amplitude versus tolerance bound. Paired difference
to the bare policy under the amplitude bound (gripper scaled with
the clip) and under the tolerance bound of Table~\ref{tab:main} ($\beta{=}0.5$,
$q{=}0.75$, as in Table~\ref{tab:main}). Mean $\pm$ sd over three
seeds, $\Delta$ paired per seed.}
\label{app:tab:boundrows}
\centering
\small
\begin{tabular}{lrrr}
\toprule
policy & amplitude & tolerance & change \\
\midrule
\multicolumn{4}{l}{\emph{LIBERO-90}} \\
ACT & $-2.0$ {\tiny $\pm$ 2.3} & $+1.4$ {\tiny $\pm$ 2.1} & $+3.4$ \\
ACT, language-conditioned & $-1.9$ {\tiny $\pm$ 0.6} & $-0.6$ {\tiny $\pm$ 0.1} & $+1.3$ \\
Diffusion Policy & $-1.4$ {\tiny $\pm$ 1.5} & $+1.2$ {\tiny $\pm$ 1.8} & $+2.6$ \\
QueST & $-5.2$ {\tiny $\pm$ 0.6} & $-2.1$ {\tiny $\pm$ 1.5} & $+3.1$ \\
VQ-BeT & $-1.8$ {\tiny $\pm$ 0.8} & $-1.3$ {\tiny $\pm$ 1.2} & $+0.5$ \\
Flow-matching DiT & $-2.2$ {\tiny $\pm$ 1.7} & $+1.5$ {\tiny $\pm$ 1.6} & $+3.7$ \\
\multicolumn{4}{l}{\emph{LIBERO-10}} \\
ACT & $-1.1$ {\tiny $\pm$ 2.4} & $+5.6$ {\tiny $\pm$ 2.9} & $+6.7$ \\
Diffusion Policy & $-0.2$ {\tiny $\pm$ 4.7} & $+3.3$ {\tiny $\pm$ 2.0} & $+3.5$ \\
Flow-matching DiT & $-4.5$ {\tiny $\pm$ 5.1} & $-2.1$ {\tiny $\pm$ 3.1} & $+2.4$ \\
\midrule
mean & & & $+3.0$ \\
\bottomrule
\end{tabular}
\end{table}

\section{Real-robot setup}
\label{app:robot}
The robot is a Franka Panda with a Robotiq 2F-85 gripper and a
wrist-mounted ZED Mini camera, in the DROID configuration
\citep{khazatsky2024droid}. We fine-tune $\pi_{0.5}$ on 400 teleoperated
demonstrations recorded at 20\,Hz. They cover two cylinders, each
stacked on one of four colored cubes, with 50 demonstrations per pair.
The library is built from the same demonstrations with the defaults of
\S\ref{sec:exp:setup} ($K{=}256$, $H{=}20$, $w_g{=}5$, $\beta{=}0.5$,
$q{=}0.75$).

The policy commands absolute end-effector poses. The layer therefore
converts each predicted chunk to per-step deltas in the base frame,
starting from the last commanded pose, quantizes it and integrates the
result back from the same pose. Consecutive commanded targets define the
deltas both in the library and at execution. Position and rotation
deltas are rescaled to the action statistics of LIBERO-90, where the
defaults were selected. The bare and the wrapped policy share the
checkpoint, the replanning interval and the scene configurations. An
attempt counts as a success if the cylinder rests on the named cube when
the arm retracts.

\textbf{Offline replay.} We replay held-out demonstrations through the
layer in five folds over episodes, stratified by task, with each library
built from the other four folds. Every step starts a 20-step chunk,
converted and quantized as on the robot. Unperturbed, the layer moves
the release position by a median of 0.5\,mm under the tolerance bound
and 1.5\,mm under the amplitude bound, while the demonstrators' own
placements spread by 5 to 8\,mm. To test a correction that the
demonstrations do not contain, we add to each chunk that contains the
release a horizontal offset perpendicular to the transport direction,
ramped in over 10 steps. At the release, the tolerance bound executes
82 / 81 / 71 / 49\% of offsets of 0.5 / 1 / 2 / 4\,cm, and the
amplitude bound 62 / 63 / 54 / 37\%.

\section{Validating the layer}
\label{app:extra}

\subsection{Pooled effect and equivalence}
\label{app:pooled}
Each of the 13 in-suite cells of Table~\ref{tab:main} (six LIBERO-90,
four LIBERO-10 and three RoboCasa rows) contributes its seed-averaged
paired difference. With cells as the unit, evaluation noise, seed
variation and differences between policies all enter the interval. The
mean is $+0.55$ points (sd 2.21), with a 95\% $t$-interval of
$[-0.78, +1.89]$ and a 90\% interval of $[-0.54, +1.64]$. Two one-sided
tests at $\alpha{=}0.05$ establish equivalence at the noise-floor margin
of $\pm 3$ points ($p{<}0.001$) and at any margin of 1.64 points or more
($p{=}0.018$ at $\pm 2$). LIBERO-90 alone gives $+0.01$
$[-1.63, +1.66]$. Adding QueST with its full 32-step chunks
(Appendix~\ref{app:extra:chunk}) gives $-0.45$ $[-2.94, +2.04]$, still
equivalent at $\pm 3$ ($p{=}0.023$).

\subsection{Fit-only control}
\label{app:extra:fitonly}
Fitting each predicted chunk to control points and decoding it again,
without a library, separates the representation from quantization. For
ACT this is the identity, since its head predicts control points, and
chunks shorter than $B$ are interpolated exactly. Where the fit does
change the chunk, it costs nothing. The Diffusion Policy moves from
92.2 {\tiny[90.1, 94.1]} to 91.9 {\tiny[89.6, 94.0]}, and QueST from
83.0 to 81.6 {\tiny[76.4, 86.1]}.

\subsection{Chunk length}
\label{app:extra:chunk}
The spline basis is defined on normalized time. A chunk of $H_\pi$ steps
is therefore fit on a uniform grid over the whole support, compared with
the library in the Gram metric of $H{=}20$, and decoded back at $H_\pi$
steps. In effect, it is matched to \protos{} rescaled in time by
$H_\pi/H$. Chunks shorter than $B$, which are the 8-step chunks of
OpenVLA-OFT and the 5-step chunks of VQ-BeT, are fit by minimum-norm
exact interpolation.

Rescaling alone costs little. OpenVLA-OFT (factor 0.4, all 8 steps
executed) changes by $-0.4$ and VQ-BeT (factor 0.25, 2 of 5 executed) by
$-1.3 \pm 1.2$, while the Diffusion Policy and the flow-matching policy
(factor 0.8, 8 of 16 executed) are at parity. QueST's native 32-step
chunks, of which 8 execute, cost $13.5 \pm 0.7$ points ($19.8 \pm 1.9$
under the amplitude bound). A library built from 32-step windows (stride
8, $K{=}256$, tolerances and Gram metric at $H{=}32$) removes the
rescaling, yet still costs $8.1 \pm 2.0$. Passing only the first 20
steps, which contain the 8 executed ones, reduces the cost to
$2.1 \pm 1.5$. Rescaling thus accounts for about 5 of the 13.5 points,
and the rest arises from matching the element and its correction to a
chunk whose last 24 steps never execute.

\subsection{Element removal}
\label{app:extra:detector}
We define a \proto{}'s failure enrichment as its usage share in failed
episodes divided by its share in successful ones (wrapped ACT,
LIBERO-90, seed 0). \Proto{} 149, a near-idle motion and the most used
one (5.0\% of chunks), is enriched $10\times$ under the default configuration
($K{=}256$, $\beta{=}0.5$, $q{=}0.75$).
When a \proto{} is removed, its chunks are redirected to the nearest
remaining one. Despite its enrichment, removing 149 leaves success
at 79.2 {\tiny[74.6, 83.7]} against 79.4. Removing five heavily used
\protos{} (149, 126, 10, 33 and 94, together 12\% of chunks) gives
79.3 {\tiny[74.6, 83.8]}. Without correction ($\beta{=}0$), the reduced
libraries score 60.7 and 57.6 against 58.8.

\subsection{The bound in physical units}
\label{app:physical}
LIBERO's operational-space controller maps a unit action to 5\,cm of
commanded end-effector displacement per step, or 0.5\,rad in
orientation. A deviation of Gram norm $c$ therefore changes a chunk's
commanded net displacement by at most $0.05\,\mathrm{m} \cdot \sqrt{H}\,
c$, which a constant offset attains. Under the default configuration, the caps
allow at most 14\,cm per 20-step chunk as the median over \protos{}, and
20\,cm at the 90th percentile.

The logged rollouts of wrapped ACT (seed 0, 900 episodes, 10{,}724 full
chunks) give the realized values. The log does not record the placement
$R$, so we compare under the yaw that minimizes the Gram distance, which
lower-bounds the distance under $R$. Every chunk stays within its cap
under this yaw (largest ratio 1.0004, from the decoder's clamp to
$[-1, 1]$). A chunk's commanded net displacement differs from that of
its \proto{} by a median of 7.1\,cm (90th percentile 12.5\,cm), against
39\,cm commanded per chunk. Since the arm achieves about 20\% of each
commanded displacement, these correspond to about 1.4 and 8\,cm of
executed motion.

\textbf{Distance to the demonstrations.} We compare each executed chunk
with the nearest of the 118{,}525 canonicalized LIBERO-90 demonstration
segments (bare ACT, seed 0, logged re-evaluation at 76.8\%). The chunks
of bare ACT lie at a median Gram distance of 0.50 (99th percentile 1.18,
maximum 1.74), and those of wrapped ACT at 0.41 (0.75, 1.06). Chunks from
failed episodes lie farther than those from successful ones (median 0.57
against 0.43). Under LIBERO-Plus (Table~\ref{app:tab:physical}), bare
ACT's chunks move away from the demonstrations on every axis, in
successful episodes as well, and the share that the layer would clip
rises from 86\% to between 90 and 92\%.

\begin{table}[h]
\caption{Distance of bare ACT's chunks to the demonstrations under
LIBERO-Plus (Table~\ref{tab:plus}). Distance: Gram distance of each
executed 20-step chunk to the nearest LIBERO-10 demonstration segment.
Clipped: share of chunks farther from their nearest \proto{} than its
cap, which the layer would clip ($K{=}256$ in-suite library,
$\beta{=}0.5$, $q{=}0.75$). Mean over three seeds.}
\label{app:tab:physical}
\centering
\small
\begin{tabular}{lrcccc}
\toprule
 & & \multicolumn{3}{c}{distance to nearest demonstration} & \\
\cmidrule(lr){3-5}
axis & chunks & median & 99th pct. & median, successes & clipped (\%) \\
\midrule
in distribution & 1870 & 0.55 & 1.50 & 0.47 & 85.6 {\tiny$\pm$ 2.3} \\
camera viewpoint & 2891 & 0.71 & 1.62 & 0.60 & 91.3 {\tiny$\pm$ 1.9} \\
robot initial state & 2304 & 0.67 & 1.60 & 0.51 & 90.4 {\tiny$\pm$ 0.9} \\
sensor noise & 2679 & 0.73 & 1.67 & 0.53 & 92.1 {\tiny$\pm$ 1.7} \\
object layout & 2240 & 0.63 & 1.59 & 0.51 & 90.0 {\tiny$\pm$ 0.5} \\
background texture & 2545 & 0.68 & 1.68 & 0.54 & 90.6 {\tiny$\pm$ 1.2} \\
lighting & 2453 & 0.67 & 1.56 & 0.51 & 89.9 {\tiny$\pm$ 1.2} \\
\bottomrule
\end{tabular}
\end{table}

\subsection{Trajectory quality}
\label{app:extra:smooth}
The LIBERO-90 rollouts of ACT, the Diffusion Policy and the
flow-matching policy under the default configuration log the executed action and
the end-effector pose at every step. For each episode we compute the
spectral arc length (SPARC) of the end-effector speed profile
\citep{balasubramanian2015sparc} and the mean action discontinuity
$\|a_t - a_{t-1}\|$, both at replanning seams and inside chunks. Failed
episodes are rougher than successful ones on every measure, so the
paired statistic uses only episodes in which both the bare and the
wrapped policy succeed (Table~\ref{tab:smooth}). No wrapped policy
differs from its bare counterpart in SPARC. Jumps inside chunks shrink
for all three policies. Jumps at seams shrink for the Diffusion Policy
and the flow-matching policy and grow by 0.007 for ACT.

\begin{table}[h]
\caption{Trajectory quality on LIBERO-90 ($\beta{=}0.5$, $q{=}0.75$,
seed~0, 900 episodes per arm). Rows: mean $\pm$ sd over successful
episodes, and over 10 demonstrations per task. $\Delta$: wrapped minus
bare, paired per task and initial state over the episodes in which both
arms succeed, with 95\% bootstrap CIs. Bold: CI excludes zero. The
Diffusion Policy's fit-only arm differs from bare on no measure.}
\label{tab:smooth}
\centering
\small
\setlength{\tabcolsep}{4pt}
\begin{tabular}{lccc}
\toprule
 & SPARC & seam jump & within-chunk jump \\
\midrule
demonstrations & $-2.38$ {\tiny$\pm$ 0.39} & 0.065 {\tiny$\pm$ 0.025} & 0.064 {\tiny$\pm$ 0.016} \\
ACT, bare & $-2.41$ {\tiny$\pm$ 0.55} & 0.212 {\tiny$\pm$ 0.092} & 0.052 {\tiny$\pm$ 0.015} \\
ACT $+$MPQ & $-2.45$ {\tiny$\pm$ 0.61} & 0.220 {\tiny$\pm$ 0.091} & 0.046 {\tiny$\pm$ 0.012} \\
\quad $\Delta$ ($n{=}617$) & $-0.01$ {\tiny[$-0.05$, $+0.02$]} & $\mathbf{+0.007}$ {\tiny[$+0.000$, $+0.014$]} & $\mathbf{-0.005}$ {\tiny[$-0.006$, $-0.005$]} \\
\midrule
Diffusion Policy, bare & $-2.44$ {\tiny$\pm$ 0.53} & 0.210 {\tiny$\pm$ 0.044} & 0.054 {\tiny$\pm$ 0.014} \\
Diffusion Policy, fit only & $-2.44$ {\tiny$\pm$ 0.50} & 0.211 {\tiny$\pm$ 0.045} & 0.054 {\tiny$\pm$ 0.014} \\
Diffusion Policy $+$MPQ & $-2.41$ {\tiny$\pm$ 0.52} & 0.201 {\tiny$\pm$ 0.041} & 0.049 {\tiny$\pm$ 0.012} \\
\quad $\Delta$ ($n{=}786$) & $+0.03$ {\tiny[$-0.01$, $+0.07$]} & $\mathbf{-0.009}$ {\tiny[$-0.012$, $-0.005$]} & $\mathbf{-0.005}$ {\tiny[$-0.005$, $-0.005$]} \\
\midrule
Flow-matching DiT, bare & $-2.44$ {\tiny$\pm$ 0.63} & 0.183 {\tiny$\pm$ 0.054} & 0.058 {\tiny$\pm$ 0.015} \\
Flow-matching DiT $+$MPQ & $-2.44$ {\tiny$\pm$ 0.67} & 0.177 {\tiny$\pm$ 0.041} & 0.052 {\tiny$\pm$ 0.013} \\
\quad $\Delta$ ($n{=}688$) & $+0.02$ {\tiny[$-0.03$, $+0.06$]} & $\mathbf{-0.006}$ {\tiny[$-0.009$, $-0.002$]} & $\mathbf{-0.006}$ {\tiny[$-0.006$, $-0.005$]} \\
\bottomrule
\end{tabular}
\end{table}

\subsection{Observation shift per axis}
\label{app:extra:plus}

Table~\ref{app:tab:plus:abs} gives the per-axis success behind
Table~\ref{tab:plus} at seed 0.

\begin{table}[h]
\caption{LIBERO-Plus success (\%) per axis (seed~0, $K{=}256$ in-suite
library, $\beta{=}0.5$, $q{=}0.75$, 600-step cap). Shifted axes: the
fixed 100-variant subset of each axis. In distribution: the ten
unperturbed tasks at 10 initial states. $\Delta$ paired per variant.}
\label{app:tab:plus:abs}
\centering
\small
\setlength{\tabcolsep}{4pt}
\begin{tabular}{lccrccrccr}
\toprule
 & \multicolumn{3}{c}{Diffusion Policy} & \multicolumn{3}{c}{ACT} & \multicolumn{3}{c}{Flow-matching DiT} \\
\cmidrule(lr){2-4} \cmidrule(lr){5-7} \cmidrule(lr){8-10}
axis & bare & $+$MPQ & $\Delta$ & bare & $+$MPQ & $\Delta$ & bare & $+$MPQ & $\Delta$ \\
\midrule
in distribution & 86 & 86 & $+0$ & 67 & 71 & $+4$ & 82 & 78 & $-4$ \\
camera viewpoint & 30 & 31 & $+1$ & 8 & 12 & $+4$ & 6 & 5 & $-1$ \\
robot initial state & 47 & 44 & $-3$ & 44 & 43 & $-1$ & 46 & 38 & $-8$ \\
sensor noise & 25 & 29 & $+4$ & 20 & 28 & $+8$ & 19 & 16 & $-3$ \\
object layout & 65 & 64 & $-1$ & 44 & 50 & $+6$ & 60 & 50 & $-10$ \\
background texture & 16 & 13 & $-3$ & 19 & 28 & $+9$ & 26 & 21 & $-5$ \\
lighting & 36 & 36 & $+0$ & 32 & 44 & $+12$ & 44 & 44 & $+0$ \\
\midrule
all shifted axes (600) & 37 & 36 & $-0$ & 28 & 34 & $+6$ & 34 & 29 & $-5$ \\
\bottomrule
\end{tabular}
\end{table}

\section{Learned selectors}
\label{app:selectors}

\paragraph{Architecture and training.}
The ACT-based selectors are the ACT network of Appendix~\ref{app:training} with
the coefficient head replaced by a $K$-way choice head, a yaw head and a
residual head, all of which read the decoder's query token. The staged
selector is trained in a single run against the frozen library. Each
demonstrated chunk is fit, yaw-canonicalized and assigned to its nearest
\proto{} under Eq.~\ref{eq:invdist} ($w_g{=}5$), which gives a label
$z^\ast$ and a yaw $\theta^\ast$. The choice head is trained by
cross-entropy on $z^\ast$ with hard labels. The yaw head predicts
$(\sin\theta, \cos\theta)$, regressed onto $\theta^\ast$, together with
a log-amplitude whose target is zero. The residual head predicts a
single offset for all control points, gripper included, regressed onto
the chunk minus $P_{z^\ast}$ placed at $\theta^\ast$. The bound is not
applied in training, so $\beta$ enters only at execution. The three
losses are weighted 1, 1 and 10. All parameters train jointly from
ImageNet initialization with the ACT recipe rather than from the
reference's weights.

The selectors on the flow DiT and Diffusion Policy read the vector
that the frozen reference computes from the observation to condition its
denoiser, namely the 512-dimensional input to the DiT's adaptive layer
norms and the 227-dimensional global conditioning of Diffusion Policy's
U-Net. Both combine ResNet-18 keypoints from the two cameras, the robot
state and the task. A linear layer and two pre-norm residual MLP blocks
(width 512, hidden 2560) map this vector to a readout, from which a
$K$-way choice head selects the element as in the staged selector.

Placement and correction, that is the yaw, the log-amplitude and the
residual of each control point, are then generated given an embedding of
the chosen element, as the reference generates actions. On the DiT, a
DiT-style transformer (width 256, five layers, adaptive layer norm on
the readout, the element and the flow time) attends over one placement
token and one token per control point and is trained by rectified flow,
with 10 Euler steps at execution. On Diffusion Policy, the placement is
a single vector rather than a sequence, so a diffusion MLP (three layers
of width 512) samples it first. Diffusion Policy's conditional 1D U-Net,
with down dimensions 104 and 208 instead of 512, 1024 and 2048, then
denoises the correction's control points, with FiLM conditioning on the
readout, the element and the sampled placement. Both are trained as
DDPMs with 100 steps and sampled with 10 DDIM steps, without sample
clipping since their targets are standardized.

The readouts and choice heads have 5.7M and 5.5M parameters and the
generators 6.3M and 6.5M, so both selectors total 12.0M. A full-width
variant of Diffusion Policy's selector uses Diffusion Policy's first two
levels (512 and 1024), since a third does not fit the ten control points,
and has a 72.7M generator. All selectors are trained for 50k steps
(batch 128, learning rate $10^{-4}$) on the reference's own 16-step
chunks, with every 50th demonstration per task held out. Every sampled
correction passes through the bound of Eq.~\ref{eq:clip}.

\paragraph{Agreement and yaw error.}
\label{app:agreement}
Agreement is the fraction of chunks whose top choice is the labeled
\proto{} $z^\ast$. We train the staged selector (seed 0) with every
50th demonstration held out. It agrees on 0.96 of held-out timesteps of
its training demonstrations but on only 0.30 of the 11{,}653 chunks of
unseen demonstrations (0.66 within its top five). On these chunks, the
predicted yaw differs from $\theta^\ast$ by a median of 0.19\,rad
($11^\circ$), and by 0.93\,rad ($53^\circ$) at the 90th percentile.

\paragraph{Frozen and end-to-end encoder.}
\label{app:frozenenc}
The frozen ACT selector of Table~\ref{tab:decomp} starts from each
seed's bare ACT, freezes the ResNet trunk, the embeddings and the
transformer encoder, and trains only the one-layer decoder, its query
and the three heads (5.6M parameters, same data and budget as the
staged selector). Offline it fits held-out timesteps slightly worse
than the staged selector (action MSE 0.013 against 0.012), yet its
success is $3.6 \pm 3.9$~points higher, within seed noise. The two
differ mainly in execution (Tables~\ref{app:tab:policies}
and~\ref{app:tab:boot}), so training the encoder on the selection
objective changes the decomposition little.

\paragraph{Rank of the selectors' choices.}
In closed loop we log the rank of each executed choice in the
reference's ordering, where rank~1 is the \proto{} nearest to the
reference's chunk. Table~\ref{app:tab:ranks} reports these ranks under
the reference's placement and correction ($L_{\mathrm{idx}}$), together
with a rank-matched rule. The rule draws
each chunk's rank independently from the selector's per-chunk ranks at
the same seed and budget and executes the \proto{} at that rank. The
staged selector's ranks have a long tail, with 28 to 29\% of its
choices beyond rank~20 and a 90th percentile of 112 to 140.
Table~\ref{app:tab:forced} reports the forced ranks of
Fig.~\ref{fig:selection}. Without any correction ($\beta{=}0$), the
staged selector's choices lie farther from the reference's chunk, yet
they cost far less than the fifth-nearest \proto{}.

\begin{table}[h]
\caption{Rank of the selectors' choices in the reference's ordering,
and the rank-matched rule (LIBERO-90, $K{=}256$). $\mathrm{SR}(L_{\mathrm{idx}})$
is re-evaluated with the ranks logged and lies within repeat noise of
Table~\ref{app:tab:policies}. The rank-matched rule executes \protos{} at ranks
drawn from the selector's own rank distribution. Mean $\pm$ sd over
three seeds, differences paired per seed. --: not evaluated.}
\label{app:tab:ranks}
\centering
\footnotesize
\setlength{\tabcolsep}{4pt}
\begin{tabular}{llcc}
\toprule
 & & \multicolumn{2}{c}{ACT selector} \\
\cmidrule(lr){3-4}
 & & staged & frozen \\
\midrule
$\beta{=}0.5$ & choices at rank 1 [\%] & 26 & 75 \\
 & choices in top 5 [\%] & 52 & 92 \\
 & median rank & 4--5 & 1 \\
 & $\mathrm{SR}(L_{\mathrm{idx}})$ & 78.7 {\tiny $\pm$ 1.9} & 78.3 {\tiny $\pm$ 2.0} \\
 & rank-matched rule & 45.0 {\tiny $\pm$ 3.2} & 71.2 {\tiny $\pm$ 1.8} \\
 & difference & $-33.6$ {\tiny $\pm$ 1.9} & $-7.1$ {\tiny $\pm$ 2.0} \\
\midrule
$\beta{=}0$ & choices at rank 1 [\%] & 19 & -- \\
 & $\mathrm{SR}_{\mathrm{ref}}$ & 59.2 {\tiny $\pm$ 1.0} & -- \\
 & $\mathrm{SR}(L_{\mathrm{idx}})$ & 56.3 {\tiny $\pm$ 1.0} & -- \\
 & $\mathrm{SR}(L)$ & 56.0 {\tiny $\pm$ 2.6} & -- \\
 & rank-matched rule $-$ $\mathrm{SR}(L_{\mathrm{idx}})$ & $-41.7$ {\tiny $\pm$ 2.0} & -- \\
\bottomrule
\end{tabular}
\end{table}

\begin{table}[h]
\caption{Forced choices under the reference's placement and
correction. Change in success when the layer executes the $r$-th
nearest \proto{} to the reference's chunk, or a uniformly random one,
instead of the nearest (LIBERO-90, $K{=}256$). Mean $\pm$ sd over three
seeds, paired per seed.}
\label{app:tab:forced}
\centering
\footnotesize
\setlength{\tabcolsep}{2.3pt}
\begin{tabular}{lccccc@{\hskip 6pt}ccc}
\toprule
 & \multicolumn{5}{c}{$\beta{=}0.5$} & \multicolumn{3}{c}{$\beta{=}0$} \\
\cmidrule(lr){2-6} \cmidrule(lr){7-9}
reference & $\mathrm{SR}_{\mathrm{ref}}$ & $r{=}2$ & $r{=}5$ & $r{=}20$ & random & $\mathrm{SR}_{\mathrm{ref}}$ & $r{=}5$ & random \\
\midrule
ACT & 80.2 {\tiny $\pm$ 1.0} & $-6.2$ {\tiny $\pm$ 0.5} & $-18.9$ {\tiny $\pm$ 2.6} & $-54.9$ {\tiny $\pm$ 1.5} & $-68.7$ {\tiny $\pm$ 1.7} & 59.2 {\tiny $\pm$ 1.0} & $-27.4$ {\tiny $\pm$ 0.6} & $-57.7$ {\tiny $\pm$ 1.0} \\
flow DiT & 83.4 {\tiny $\pm$ 1.4} & $-4.6$ {\tiny $\pm$ 1.1} & $-12.7$ {\tiny $\pm$ 1.5} & $-47.4$ {\tiny $\pm$ 3.3} & $-66.3$ {\tiny $\pm$ 1.2} & 62.6 {\tiny $\pm$ 1.9} & $-25.0$ {\tiny $\pm$ 0.8} & $-60.2$ {\tiny $\pm$ 1.6} \\
Diff.\ Policy & 92.9 {\tiny $\pm$ 1.2} & $-2.6$ {\tiny $\pm$ 0.9} & $-10.4$ {\tiny $\pm$ 0.7} & $-50.2$ {\tiny $\pm$ 2.3} & $-73.1$ {\tiny $\pm$ 1.3} & 74.9 {\tiny $\pm$ 1.8} & $-24.5$ {\tiny $\pm$ 3.4} & $-71.9$ {\tiny $\pm$ 2.2} \\
\bottomrule
\end{tabular}
\end{table}

\paragraph{Element-conditioned heads.}
The control replaces the staged selector's linear yaw and residual
heads by two-layer MLPs (width 512, GELU, 563k parameters against 37k).
The conditioned variant adds a linear embedding of the chosen
\proto{}'s control points to the decoder token (36k parameters,
$+0.10\%$ of the model). It is trained on the label $z^\ast$ and
executes the selector's own choice. The third variant has the same
architecture, but after a 10k-step warm-up it conditions on the
selector's own detached choice, and its residual target is computed
against that element. All variants use the recipe and bound of the
staged selector. Table~\ref{app:tab:cond} reports their
decomposition. Relative to the control, the execution term changes by
$+2.2 \pm 3.4$ with conditioning and $+3.0 \pm 3.6$ with supervision on
the own choice. At seed~0 the bound clips 62\% and 59\% of the
conditioned heads' chunks, against 34\% for the control, so their gain
does not come from a looser bound.

\begin{table}[h]
\caption{Selector heads with and without the chosen element
(ACT staged, LIBERO-90, $K{=}256$, $\beta{=}0.5$). Execution in both
substitution orders. Cosine: median cosine in the Gram metric between
the learned correction and the demonstrated one on held-out frames
(ACT's own correction: 0.74). Mean $\pm$ sd over three seeds, paired
per seed.}
\label{app:tab:cond}
\centering
\footnotesize
\setlength{\tabcolsep}{4pt}
\begin{tabular}{lccccc}
\toprule
 & & & \multicolumn{2}{c}{execution} & \\
\cmidrule(lr){4-5}
heads & $\mathrm{SR}(L)$ & $\mathrm{SR}(L_{\mathrm{idx}})$ & choice first & correction first & cosine \\
\midrule
linear (staged) & 63.6 {\tiny $\pm$ 1.7} & 78.1 {\tiny $\pm$ 0.6} & $-14.6$ {\tiny $\pm$ 1.5} & $-19.0$ {\tiny $\pm$ 0.4} & 0.47 \\
MLP control & 65.6 {\tiny $\pm$ 1.5} & 77.5 {\tiny $\pm$ 1.2} & $-11.9$ {\tiny $\pm$ 2.1} & $-20.6$ {\tiny $\pm$ 2.2} & 0.52 \\
conditioned, trained on label & 69.7 {\tiny $\pm$ 1.6} & 79.4 {\tiny $\pm$ 0.8} & $-9.7$ {\tiny $\pm$ 1.4} & $-10.9$ {\tiny $\pm$ 0.5} & 0.66 \\
conditioned, trained on own choice & 69.9 {\tiny $\pm$ 1.1} & 78.9 {\tiny $\pm$ 0.5} & $-9.0$ {\tiny $\pm$ 1.6} & $-12.7$ {\tiny $\pm$ 1.2} & 0.65 \\
\bottomrule
\end{tabular}
\end{table}

\paragraph{Offline diagnosis of the learned correction.}
\label{app:resid}
We compare corrections on each seed's validation split, a random 1/50 of
timesteps (11{,}580 frames) that ACT and the selector both hold out.
Each frame is executed with the evaluator's code at $\beta{=}0.5$. For
the selector's element placed at the demonstrated yaw, the target
$e^\ast$ is the demonstrated chunk minus that element. We report medians
over frames, averaged over seeds (seed sd at most 0.01). The linear
heads' correction is poorly aligned with $e^\ast$ (median cosine 0.47,
Table~\ref{app:tab:cond}) and is shorter, at 0.43 of $\|e^\ast\|_M$ against 0.90 for ACT's.
Direction matters more than length. The executed chunk lies at a median
Gram distance of 0.943 from the demonstration, 0.964 with the learned
correction rescaled to the demonstrated norm and 0.616 with the
demonstrated direction at the learned norm, against 0.450 for $e^\ast$
itself. The selector's yaw error matches ACT's (median $6.4^\circ$
against $6.5^\circ$), so the difference lies in the correction. These
offline errors do not order the selectors by closed-loop success. The
frozen-encoder selector has the largest offline error (0.980) but
scores above the linear heads in closed loop.

\paragraph{Order of the terms.}
The hybrid $L_{\mathrm{exec}}$ executes the \proto{} nearest to the
reference's chunk, placed by the selector's yaw head and corrected by
its residual head under the same bound. Table~\ref{app:tab:policies}
gives the success of every executed policy, Table~\ref{tab:decomp} the
terms in both orders and Table~\ref{app:tab:boot} their bootstrap
intervals and the order-averaged selection term.
Averaged over both orders, execution changes success by
$-16.8 \pm 0.9$, $-12.3 \pm 1.7$, $-1.6 \pm 1.2$ and $-9.7 \pm 0.3$
for the staged selector and the three selectors of
Table~\ref{tab:decomp}.

\begin{table}[h]
\caption{Success of the executed policies behind
Table~\ref{tab:decomp}, of the staged selector and of Diffusion
Policy's full-width selector (LIBERO-90, $K{=}256$, $\beta{=}0.5$). Mean $\pm$
sd over three seeds.}
\label{app:tab:policies}
\centering
\footnotesize
\setlength{\tabcolsep}{2.5pt}
\begin{tabular}{llrrrrr}
\toprule
 & & staged & \multicolumn{4}{c}{frozen encoder} \\
\cmidrule(lr){3-3} \cmidrule(lr){4-7}
 & executed policy & ACT & ACT & flow DiT & \multicolumn{2}{c}{Diff.\ Policy} \\
 & heads & regression & regression & flow & diffusion & full width \\
\midrule
$\mathrm{SR}(\pi)$ & reference, bare
  & \multicolumn{2}{c}{78.8 {\tiny $\pm$ 2.3}} & 81.9 {\tiny $\pm$ 1.5} & \multicolumn{2}{c}{91.7 {\tiny $\pm$ 0.6}} \\
$\mathrm{SR}_{\mathrm{ref}}$ & reference through the library
  & \multicolumn{2}{c}{80.2 {\tiny $\pm$ 1.0}} & 83.4 {\tiny $\pm$ 1.4} & \multicolumn{2}{c}{92.9 {\tiny $\pm$ 1.2}} \\
$\mathrm{SR}(L_{\mathrm{idx}})$ & selector's choice, reference's correction
  & 78.1 {\tiny $\pm$ 0.6} & 79.3 {\tiny $\pm$ 2.0} & 83.0 {\tiny $\pm$ 2.0} & 92.4 {\tiny $\pm$ 0.1} & 92.0 {\tiny $\pm$ 0.8} \\
$\mathrm{SR}(L_{\mathrm{exec}})$ & reference's choice, selector's correction
  & 61.2 {\tiny $\pm$ 0.9} & 67.7 {\tiny $\pm$ 3.1} & 80.3 {\tiny $\pm$ 1.2} & 83.8 {\tiny $\pm$ 1.0} & 87.6 {\tiny $\pm$ 1.2} \\
$\mathrm{SR}(L)$ & selector's choice and correction
  & 63.6 {\tiny $\pm$ 1.7} & 67.1 {\tiny $\pm$ 3.5} & 83.0 {\tiny $\pm$ 0.0} & 82.0 {\tiny $\pm$ 2.0} & 85.8 {\tiny $\pm$ 2.1} \\
\bottomrule
\end{tabular}
\end{table}

\begin{table}[h]
\caption{Terms of Table~\ref{tab:decomp}, of the staged selector and of
Diffusion Policy's full-width selector, with paired 95\% percentile
bootstrap intervals over the 90 tasks ($B{=}10{,}000$, per-task
differences averaged over the three seeds). Intervals that also resample
seeds are up to 2.6~points wider. They still exclude zero for every
execution term of the ACT selectors, but not for the staged selector's
single-order selection terms.}
\label{app:tab:boot}
\centering
\small
\begin{tabular}{lccccc}
\toprule
 & \multicolumn{2}{c}{execution} & \multicolumn{3}{c}{selection} \\
\cmidrule(lr){2-3}\cmidrule(lr){4-6}
 & choice & correction & choice & correction & order \\
selector & first & first & first & first & average \\
\midrule
ACT staged & $-14.6$ & $-19.0$ & $-2.1$ & $+2.3$ & $+0.1$ \\
 & {\scriptsize [$-17.7$, $-11.4$]} & {\scriptsize [$-21.7$, $-16.3$]} & {\scriptsize [$-3.9$, $-0.2$]} & {\scriptsize [$+0.1$, $+4.6$]} & {\scriptsize [$-1.4$, $+1.8$]} \\
\addlinespace
ACT frozen & $-12.1$ & $-12.4$ & $-0.9$ & $-0.6$ & $-0.8$ \\
 & {\scriptsize [$-14.7$, $-9.7$]} & {\scriptsize [$-15.1$, $-9.9$]} & {\scriptsize [$-2.3$, $+0.5$]} & {\scriptsize [$-2.8$, $+1.5$]} & {\scriptsize [$-2.1$, $+0.6$]} \\
\addlinespace
flow DiT & $+0.0$ & $-3.1$ & $-0.4$ & $+2.7$ & $+1.2$ \\
 & {\scriptsize [$-1.9$, $+1.8$]} & {\scriptsize [$-4.6$, $-1.6$]} & {\scriptsize [$-2.0$, $+1.2$]} & {\scriptsize [$+0.9$, $+4.5$]} & {\scriptsize [$-0.1$, $+2.5$]} \\
\addlinespace
Diff.\ Policy & $-10.3$ & $-9.0$ & $-0.5$ & $-1.8$ & $-1.1$ \\
 & {\scriptsize [$-12.4$, $-8.4$]} & {\scriptsize [$-11.3$, $-6.9$]} & {\scriptsize [$-1.7$, $+0.7$]} & {\scriptsize [$-3.6$, $-0.0$]} & {\scriptsize [$-2.2$, $-0.1$]} \\
\addlinespace
Diff.\ Policy, full width & $-6.3$ & $-5.2$ & $-0.8$ & $-1.9$ & $-1.3$ \\
 & {\scriptsize [$-8.0$, $-4.7$]} & {\scriptsize [$-7.1$, $-3.4$]} & {\scriptsize [$-2.0$, $+0.3$]} & {\scriptsize [$-3.4$, $-0.3$]} & {\scriptsize [$-2.4$, $-0.3$]} \\
\bottomrule
\end{tabular}
\end{table}

\paragraph{Other selector designs.}
The co-adapted selectors train the \protos{} jointly with the selector
under the IMC curriculum \citep{blessing2023imc}, which enforces a
target fraction of \protos{} in use (0.95 and 0.70). The mixture
selector instead executes a trained sparse mixture of its top-2
\protos{}. All three start from the seed-0 $w_g{=}1$, $K{=}256$
library, whose library reference is within noise of the default
(Table~\ref{app:tab:designs}). They score 40.7 {\tiny[34.0, 47.4]}, 29.4
{\tiny[23.6, 35.6]} and 28.9 {\tiny[23.1, 34.9]}, against 61.6 for the
staged selector at seed 0. All three saturate the bound on every chunk,
with a median residual of 5 to 8 times the bound.

\paragraph{Language conditioning.}
\label{app:language}
Conditioned on CLIP embeddings of the instruction instead of the task
id, bare ACT scores $72.1 \pm 0.8$ and wrapped ACT $71.6 \pm 0.8$, a
paired cost of $0.6 \pm 0.1$ (Table~\ref{app:tab:main}). A
language-conditioned staged selector, on a $K{=}64$ library
(yaw and amplitude canonicalization, $w_g{=}1$), scores
45.3 {\tiny[39.7, 51.1]}, 27 points below bare, against 15 under
task-id conditioning. Loosening its bound to $\beta{=}2$, where almost
no chunk is clipped, gives 47.7 {\tiny[42.2, 53.2]}, and removing its
residual ($\beta{=}0$) gives 32.1. A learned residual freed of the bound
thus does not recover the loss.

\FloatBarrier
\section{Library reference of a learned codebook}
\label{app:extra:codec}
The library reference applies to a latent codebook only in a confounded form.
QueST's codec consumes 32-step windows, so a reference's 20-step chunks
must first be resampled. The resampling alone lowers ACT from 79.9 to
28.2 and the Diffusion Policy from 91.7 to 59.6. On identical initial
states, the codec round trip then changes the resampled ACT by $+1.7$
$[-2.4, +5.8]$ and the resampled Diffusion Policy by $-10.3$
$[-14.3, -6.3]$. Because the decoder and its fixed time support cannot
be separated from the quantization, Eq.~\ref{eq:decomp} is ill-posed for
this codebook. A tokenizer without these confounds behaves like the
library. Round-tripping every Diffusion Policy chunk through FAST
\citep{pertsch2025fast}, which encodes each chunk at its native length,
gives 92.8 {\tiny[90.3, 95.0]} against 92.2 bare.

\section{Library design, size, transfer and editing}
\label{app:extra:offline}

\subsection{Design sweep}
\label{app:design}
Table~\ref{app:tab:designs} gives the full design sweep, and
Table~\ref{tab:ablations} is its $K{=}256$, $w_g{=}5$ column. Yaw
canonicalization adds 10 points at $K{=}64$ and 3 to 4 at $K{=}256$, so it
matters most where the library is too small to absorb the variety of
orientations. The Gram metric adds 3 to 4 points over Euclidean distance
at $K{=}64$ and is within noise at $K{=}256$. The two gripper weights
are within noise of each other at $K{=}256$. Unprojected centroids score
79.8 {\tiny[75.2, 84.0]} against 79.4 for projected ones, and an
IMC-style curriculum mixture \citep{blessing2023imc,li2023curriculum}
scores 77.8 {\tiny[72.4, 82.7]}. The number of control points has
little effect, with $B{=}6$, 10 and 14 giving 93.0, 92.1 and 93.4 for
the Diffusion Policy.

\begin{table}[h]
\caption{Closed-loop library reference of every library design. ACT (seed~0, bare
79.9) executed through each library at $\beta{=}0.5$, $q{=}0.75$, one
900-rollout run per cell (noise floor about 3 points). Rows form a
cumulative ladder. Euclidean: clustering and assignment by Euclidean
distance on control points. Gram: by the metric of Eq.~\ref{eq:gram},
without canonicalization. $w_g$: gripper-channel weight.}
\label{app:tab:designs}
\centering
\small
\begin{tabular}{lcccc}
\toprule
 & \multicolumn{2}{c}{$K{=}64$} & \multicolumn{2}{c}{$K{=}256$} \\
\cmidrule(lr){2-3} \cmidrule(lr){4-5}
design & $w_g{=}1$ & $w_g{=}5$ & $w_g{=}1$ & $w_g{=}5$ \\
\midrule
Euclidean & 65.6 & 62.0 & 73.2 & 74.6 \\
Gram & 69.1 & 66.0 & 76.4 & 76.7 \\
\quad $+$ yaw (ours) & 79.3 & 76.0 & 80.3 & 79.4 \\
\quad $+$ amplitude & 73.9 & 73.0 & 78.4 & 74.4 \\
\bottomrule
\end{tabular}
\end{table}

\subsection{Library size on RoboCasa}
Doubling the in-suite RoboCasa library to $K{=}512$ changes no policy
by more than a point against bare. ACT changes by $+0.4 \pm 2.7$ and
the Diffusion Policy by $+0.6 \pm 1.9$ (three seeds), and the
flow-matching policy by $-0.6$ {\tiny[$-5.1$, $+4.1$]} (seed 0). Over
$K \in \{64, 128, 256, 512, 1024\}$, the Diffusion Policy scores 40.0,
40.9, 42.1, 42.7 and 44.8 (bare 43.5), with task-level intervals of
about $\pm 11$.

\subsection{Transfer}
The LIBERO-90 library explains 89.0\% of the variance of the
402{,}344 RoboCasa arm segments, against 92.0\% for RoboCasa's own
library. On LIBERO-10 it executes ACT at $64.5 \pm 2.9$, at parity with
bare ACT and $5.8 \pm 2.6$ below LIBERO-10's own library. The occupancy
counts of \S\ref{sec:libraries:transfer} are effective numbers of
\protos{}, the exponential of the entropy of a corpus's assignment
frequencies.

\subsection{Editing}
Table~\ref{tab:edit} reports edits of the LIBERO-90 library toward
RoboCasa. Re-clustering the union of the two corpora at $K{=}256$ brings
RoboCasa within 1.2 points of its own library and costs LIBERO-90
$0.4 \pm 0.6$. At $K{=}512$, neither column moves by more than a point.
Appending 64 RoboCasa \protos{} leaves LIBERO-90 unchanged and recovers
part of the RoboCasa shortfall ($-3.3$ against $-4.8$). Choosing the 64
appended \protos{} differently, as RoboCasa's own $K{=}64$ library or at
random, makes no measurable difference (seed 0, five-point intervals
over 17 tasks). For the Diffusion Policy (seed 0), the re-clustered
library gives $+1.2$ on RoboCasa and $+1.0$ on LIBERO-90. Each edit is a
two-minute clustering or a concatenation, and the policies remain
untouched.

\begin{table}[h]
\caption{Library editing toward RoboCasa (ACT, $\beta{=}0.5$,
$q{=}0.75$). Paired difference against each suite's own $K{=}256$
library. Mean $\pm$ sd over three seeds where given, otherwise seed~0
with 95\% bootstrap CIs over tasks. Appended: the 64 most-populated of
RoboCasa's 256 \protos{}.}
\label{tab:edit}
\centering
\small
\begin{tabular}{lrrr}
\toprule
 & & \multicolumn{2}{c}{$\Delta$ against own library} \\
\cmidrule(lr){3-4}
edit & $K$ & RoboCasa policy & LIBERO-90 policy \\
\midrule
none (LIBERO-90 library) & 256 & $-4.8$ {\tiny[$-10.2$, $+0.9$]} & --- \\
re-cluster LIBERO-90 $\cup$ RoboCasa & 256 & $-1.2$ {\tiny[$-5.5$, $+2.7$]} & $-0.4$ {\tiny $\pm$ 0.6} \\
 & 512 & $-2.0$ {\tiny[$-6.7$, $+2.9$]} & $-0.6$ {\tiny[$-3.4$, $+2.4$]} \\
append 64 RoboCasa prototypes & 320 & $-3.3$ {\tiny[$-6.8$, $+0.1$]} & $-0.2$ {\tiny $\pm$ 0.7} \\
\bottomrule
\end{tabular}
\end{table}

\paragraph{Deleting fast \protos{}.}
\label{app:edit:speed}
We delete from the $K{=}256$ library the 64 \protos{} whose largest
commanded per-step translation reaches 0.843 in normalized action units,
the library's 75th percentile. Table~\ref{app:tab:speed} reports wrapped
ACT with the full and the edited library, together with the share of
executed chunks whose commanded speed reaches this threshold. With the
full library, tightening the bound barely lowers this share, because
the library's own fast elements remain. The threshold is a quantile of
the library, not a physical limit, and speeds are commanded, not
measured.

\begin{table}[h]
\caption{Deleting the fastest quarter of the library (ACT,
LIBERO-90, $K{=}256$). Success: mean $\pm$ sd over three seeds. Effect
of deletion: paired against the full library at the same $\beta$, with
95\% bootstrap CIs over tasks. Fast: executed chunks with commanded
speed $\geq 0.843$. All cells were evaluated together, so bare and
full-library success differ from Table~\ref{tab:main} within repeat noise.}
\label{app:tab:speed}
\centering
\small
\setlength{\tabcolsep}{4pt}
\begin{tabular}{lcccrcc}
\toprule
 & \multicolumn{2}{c}{success} & & & \multicolumn{2}{c}{fast [\%]} \\
\cmidrule(lr){2-3} \cmidrule(lr){6-7}
$\beta$ & full & deleted & & effect of deletion & full & deleted \\
\midrule
bare ACT & \multicolumn{2}{c}{79.0 {\tiny $\pm$ 0.6}} & & & \multicolumn{2}{c}{28.4} \\
0.5 & 79.2 {\tiny $\pm$ 1.8} & 79.3 {\tiny $\pm$ 1.2} & & $+0.0$ {\tiny[$-1.6$, $+1.6$]} & 19.6 & 11.9 \\
0.25 & 71.9 {\tiny $\pm$ 1.0} & 69.1 {\tiny $\pm$ 0.6} & & $-2.7$ {\tiny[$-5.0$, $-0.4$]} & 16.0 & 4.5 \\
0.1 & 64.6 {\tiny $\pm$ 0.4} & 61.0 {\tiny $\pm$ 0.3} & & $-3.5$ {\tiny[$-7.1$, $-0.2$]} & 14.4 & 0.8 \\
0 & 58.5 {\tiny $\pm$ 1.7} & 53.7 {\tiny $\pm$ 2.1} & & $-4.9$ {\tiny[$-8.0$, $-1.8$]} & 13.8 & 0.0 \\
\bottomrule
\end{tabular}
\end{table}

\FloatBarrier
\section{Extended related work}
\label{app:related}

\paragraph{Discrete skill representations.}
BeT \citep{shafiullah2022bet} is the closest structural relative, since
its bin plus offset resembles a \proto{} plus residual. Its bins,
however, are means of raw single-step actions rather than canonicalized
chunks under a trajectory metric. Its selector is a trained classifier
rather than a lookup on the policy's own output, and its offsets are
unbounded. Methods that learn continuous skill latents
\citep{pertsch2020spirl,ajay2021opal,nasiriany2022sailor} place the
bottleneck inside the policy but have no quantization term. Concurrent
work pairs a learned selector with an unbounded residual
\citep{lei2026pfdag,wang2026arp}. In our experiments, an unbounded
regression residual does not recover what a selector's learned
correction loses, while a generative head recovers part of it within
the bound (\S\ref{sec:placement}, Appendix~\ref{app:language}).

\paragraph{Trajectory primitives.}
Dynamical movement primitives encode motions as attractor systems
\citep{ijspeert2013dynamical}, and probabilistic movement primitives
encode them as distributions over trajectories
\citep{paraschos2013probabilistic,li2023prodmp}. Trajectory libraries
turn stored trajectories into a policy by executing the one whose states
are nearest to the current state \citep{stolle2006trajectory}. Maneuver
automata compose a finite set of maneuvers, each defined up to the
symmetry group of the system, so that one stored maneuver serves every
translated or rotated instance \citep{frazzoli2005maneuver}. The yaw
canonicalization of \S\ref{sec:method:canon} applies the same idea to
demonstration segments, restricted to rotations about gravity.

\paragraph{Prototypes, retrieval and runtime projection.}
ProtoPNet \citep{chen2019protopnet} classifies inputs by their
similarity to prototypes projected onto training examples, and it has
been extended to sequences \citep{ming2019prosenet}.
We adopt its projection step, and the selector of \S\ref{sec:placement}
is a classifier over the library's elements. Retrieval of
sub-trajectories for training \citep{memmel2025strap,nasiriany2022sailor}
leaves the executed action unconstrained, as retrieval over observations
does. Bidirectional decoding \citep{liu2025bid} samples several chunks
and keeps the one most consistent with past and likely future actions,
so the executed chunk is still one that the policy produced. Residual
policy learning \citep{silver2018residual,johannink2019residual} trains a
correction on top of a fixed controller. MPQ inverts these roles,
since the demonstrated segment is fixed and the policy's prediction
supplies a bounded correction. Safety shields and
control-barrier-function filters
\citep{alshiekh2018shielding,ames2019cbf} project a
policy's action onto an implicit safe set rather than onto a finite
library, and runtime monitors \citep{ahmed2026tailsafe} detect unsafe
executions without changing the action.

\section{Prototype cards}
\label{app:cards}

\begin{figure}[h]
\centering
\includegraphics[width=\linewidth]{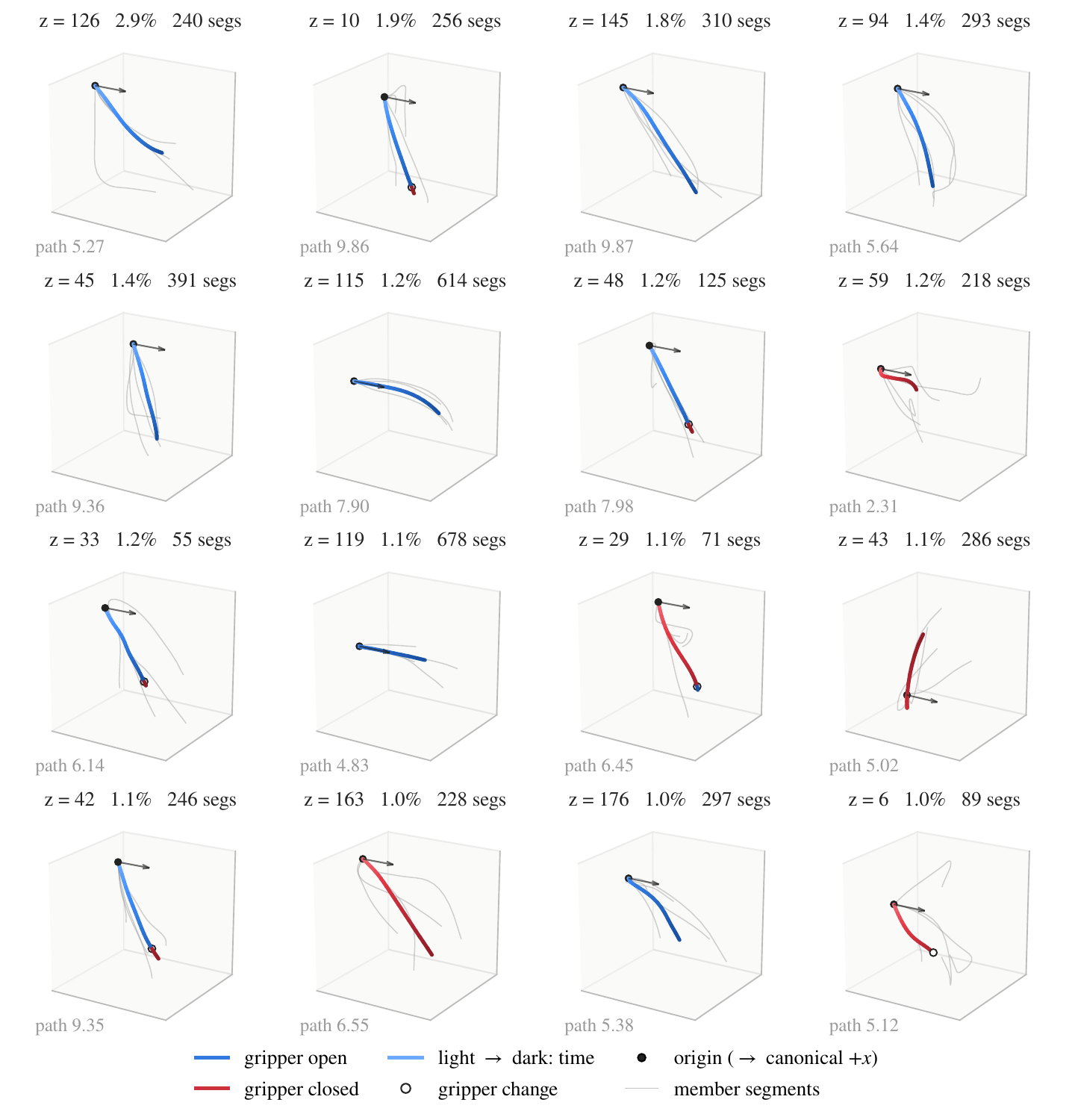}
\caption{The sixteen most-used \protos{} of wrapped ACT on LIBERO-90,
ranked by closed-loop usage over the rollouts of \S\ref{sec:exp:main},
excluding the near-idle $z{=}149$ (Fig.~\ref{fig:protos},
Appendix~\ref{app:extra:detector}). Each card shows the decoded
end-effector path in the canonical frame from a shared camera, scaled to
its own extent. Headers: index $z$, usage share and number of member
segments. Path: path length in action units, which restores absolute
scale.}
\label{fig:cards}
\end{figure}

\end{document}